\documentclass[10pt,twocolumn,letterpaper]{article}

\usepackage{cvpr}
\usepackage{times}
\usepackage{epsfig}
\usepackage{graphicx}
\usepackage{amsmath}
\usepackage{amssymb}
\usepackage{multirow}

\usepackage[pagebackref=true,breaklinks=true,letterpaper=true,colorlinks,bookmarks=false]{hyperref}

\newcommand{\bd}[1]{\textbf{#1}}
\newlength\savewidth

\cvprfinalcopy 

\ifcvprfinal\fi
\begin{document}

\title{Multi-Task Multi-Sensor Fusion for 3D Object Detection}

\author{
  Ming Liang$^{1}$\thanks{Equal contribution.} \quad Bin Yang$^{1,2 *}$ \quad Yun Chen$^{1}$\thanks{Work done as part of Uber AI Residency program (\url{https://careersinfo.uber.com/ai-residency}).} \quad Rui Hu$^{1}$ \quad Raquel Urtasun$^{1,2}$\\
  $^{1}$Uber Advanced Technologies Group \quad $^{2}$University of Toronto\\
  \small\texttt{\{ming.liang, byang10, yun.chen, rui.hu, urtasun\}@uber.com}
}

\maketitle

\begin{abstract}

In this paper we propose to exploit multiple related tasks for accurate  multi-sensor 3D object detection. Towards this goal we present an end-to-end learnable architecture that reasons about 2D and 3D object detection as well as ground estimation and depth completion. 
Our experiments show that all these tasks are complementary  and help the network learn better representations by fusing information at various levels. 
Importantly, our approach leads the KITTI benchmark on 2D, 3D and BEV object detection, while being real time.  

\end{abstract}
\section{Introduction}

Self driving vehicles have the potential to  improve safety, provide mobility solutions for otherwise underserved sectors of the population and reduce pollution.  Fundamental to its core is the ability to perceive the scene in real-time. Most autonomous driving systems rely on 3-dimensional perception, as it enables interpretable motion planning in bird's eye view.   

Over the past few years we have seen a plethora of methods that tackle the problem of 3D object detection from  monocular images \cite{mono3d, mono3d18}, stereo cameras \cite{3doppami} or LiDAR point clouds \cite{voxelnet, pixor, dpt}. 
However, each sensor has its challenges: cameras have difficulty capturing fine-grained 3D information, while LiDAR provides very sparse observations at long range.
Recently, several attempts \cite{mv3d, fpointnet, avod, contfuse} have been developed to fuse information from multiple sensors. 
Methods like \cite{fpointnet,fpccnn} adopt a cascade approach by using cameras in the first stage and reasoning using point clouds  from LiDAR-only at the second stage. However, such cascade approach  suffers from the weakness of each single sensor. As a result, it is difficult to detect objects that are occluded or far away. Others \cite{mv3d, avod, contfuse} have proposed to fuse features instead. 
Single-stage detectors like \cite{contfuse} fuse multi-sensor feature maps using LiDAR point as pixel correspondence. Local nearest neighbor interpolation is used to densify the correspondence. However, the fusion is limited when LiDAR points become extremely sparse at long range.
Two-stage detectors  \cite{mv3d, avod} fuse multi-sensor features per object at Region-Of-Interest (ROI) level. However, the fusion process is slow (as it involves thousands of ROIs) and imprecise (either using fix-sized anchors or ignoring object orientation).

\begin{figure}[!t]
\begin{center}
 \includegraphics[width=1.0\linewidth]{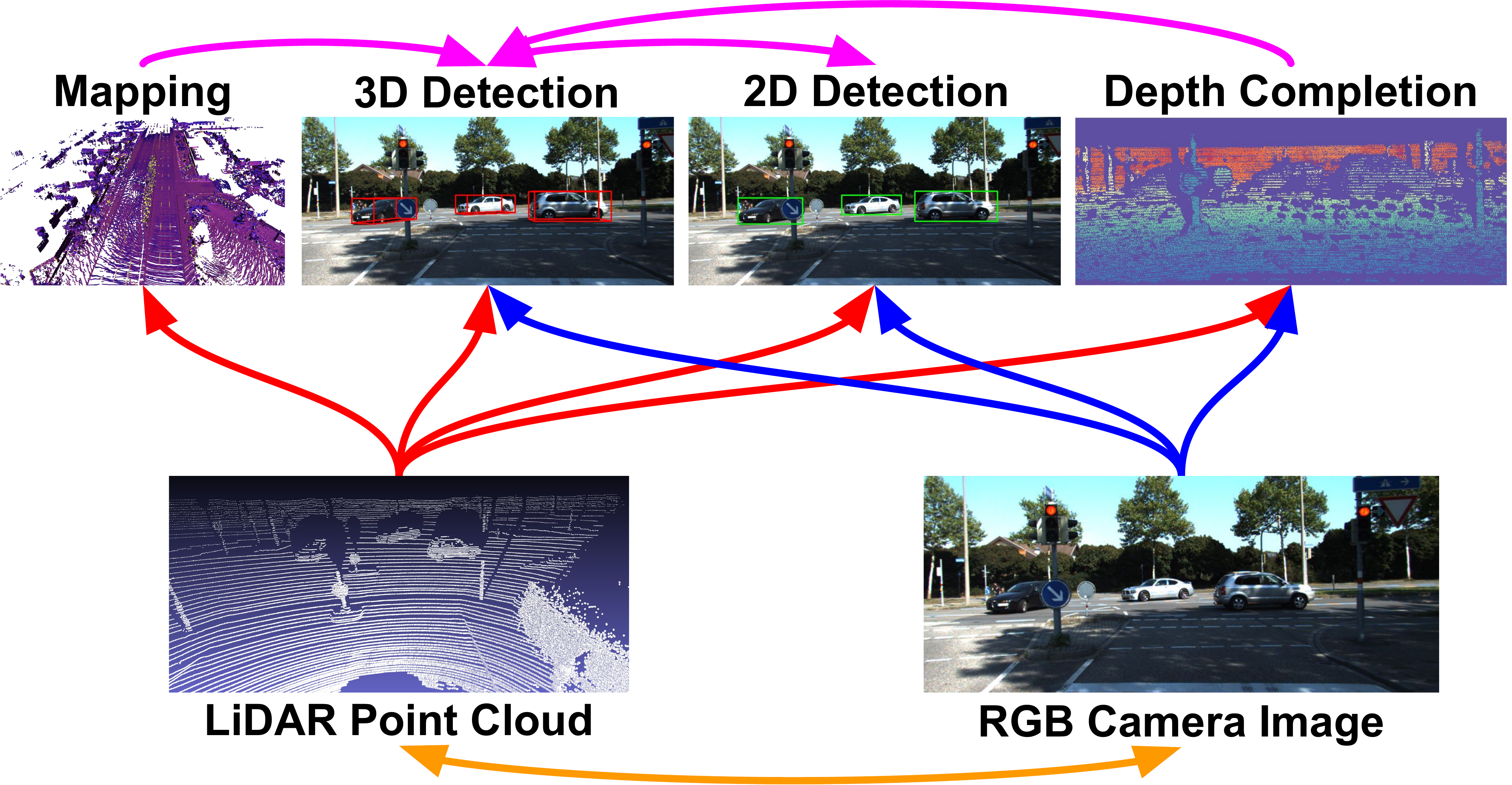}
\end{center}
   \caption{Different sensors (bottom) and tasks (top) are complementary to each other. We propose a joint model that reasons on two sensors and four tasks, and show that the target task - 3D object detection can benefit from multi-task learning and multi-sensor fusion.}
\label{fig:trailer}
\end{figure}

In this paper we argue that by performing multiple perception tasks, we can learn better feature representations that result in better detection performance. 
Towards this goal, we developed a multi-sensor
detector that reasons about 2D and 3D object detection, ground estimation and depth completion. 
Importantly, our model can be learned end-to-end and performs all these tasks at once. 
We refer the reader to Fig. \ref{fig:trailer} for an illustration of our approach.

We propose a new multi-sensor fusion architecture that leverages the advantages from both point-wise and ROI-wise feature fusion, resulting in  fully fused feature representations. 
Knowledge about the location of the ground can provide useful cues for 3D object detection in the context of self driving, as the traffic participants stick out of it. Our detector estimates an accurate pointwise  ground location online as one of its auxiliary tasks. This in turn is used by the main bird's eye view (BEV) backbone to  reason about relative location. 
We also exploit the task of depth completion to learn better cross-modality feature representation and more importantly, help achieve dense point-wise feature fusion.

We demonstrate the effectiveness of our approach on the KITTI object detection benchmark \cite{kitti} as well as the more challenging TOR4D object detection benchmark \cite{pixor}. On the KITTI benchmark, we show very significant performance improvement over other state-of-the-art approaches in 2D, 3D and Bird's Eye View (BEV) detection tasks. 
In particular, we surpass the second best 3D detector by over $3\%$ in Average Precision (AP).
Meanwhile, the proposed detector also runs over 10 frames per second, making it a practical solution for real-time applications. On the TOR4D benchmark, we show detection improvement from multi-task learning over previous state-of-the-art detector.

\section{Related Work}
We focus our literature review on works that exploit multi-sensor fusion and multi-task learning to improve 3D object detection.
\newline
\newline
\bd{3D detection from single modality:}
Early approaches to 3D object detection focus on camera based solutions, with monocular or stereo images \cite{3dop, mono3d}. However, they suffer from the inherent difficulties of estimating depth from images  and as a result  perform poorly in 3D localization. More recent  3D object detectors rely on depth sensors such as  LiDAR \cite{pixor, voxelnet}. However, although range sensors provide precise depth measurements,  the observations are usually  sparse (particularly at long range) and lack the information richness of  images. It is thus difficult to distinguish classes such as pedestrian and bicyclist with LiDAR-only detectors. 
\newline
\newline
\bd{Multi-sensor fusion for 3D detection:}
Recently, a variety of 3D detectors that exploit  multiple sensors (e.g., LiDAR and camera) have been proposed. 
F-PointNet \cite{fpointnet} uses a cascade approach to fuse multiple sensors. Specifically, 2D object detection is done first on images, 3D frustums are then generated by projecting 2D detections to 3D and PointNet \cite{pointnet, pointnet2} is applied to regress the 3D position and shape of the bounding box. In this framework the overall performance is bounded by either stage which is still using single sensor. Furthermore, regressing positions from a frustum in LiDAR point cloud has difficulty dealing with occluded or far away objects as LiDAR observation can be very sparse (often containing a single point on the object). 
MV3D \cite{mv3d} generates 3D proposals from LiDAR features, and refines the detections by Region-Of-Interest (ROI) feature fusion from LiDAR and image features. AVOD \cite{avod} further adds ROI feature fusion to the proposal generation stage to improve the proposal quality. However, ROI feature fusion happens only  at high-level feature maps. Furthermore,  it only fuses features at selected object regions instead of densely over the feature map. To overcome this drawback, ContFuse \cite{contfuse}  uses continuous convolutions to fuse multi-scale convolutional feature maps, where the  correspondence between modalities is computed through projection of the LiDAR points. However, such fusion is limited when LiDAR points are very sparse. To resolve this issue, in this paper we propose to predict dense depth from LiDAR and image and use the predicted depth points to find dense correspondences between the  feature maps from the two sensor modalities. 
\newline
\newline
\bd{3D detection from multi-task learning:}
Various tasks  have been exploited to help improve 3D object detection. HDNET \cite{hdnet} exploits geometric ground shape and semantic road masks to improve 3D object detection. 
Our model also reasons about a geometric map. The difference is that this module is part of our detector and  thus end-to-end trainable, so that these two tasks can be optimized jointly. Wang et al. \cite{holistic} exploit depth reconstruction and semantic segmentation to help 3D object detection. However, they  rely on rendering, which  is computationally expensive. Other contextual cues such as the room layout \cite{rencvpr16, schwing2013box}, and support surface \cite{rencvpr18} have  also been exploited to help 3D object reasoning in the context of indoor scenes.
3DOP \cite{3dop} exploits monocular depth estimation to refine the 3D shape and position based on 2D proposals. Mono3D \cite{mono3d} proposes to use instance segmentation and semantic segmentation as evidence, along with other geometric priors to reason about 3D object detection from monocular images. 
In  contrast to the aforementioned approaches, in this paper we also exploit depth completion  which provides two benefits: it guides the network to learn better cross-modality feature representations  and its prediction is exploited for dense pixel-wise feature fusion between the two-stream backbone networks.

\begin{figure*}[t]
\begin{center}
 \includegraphics[width=1.0\linewidth]{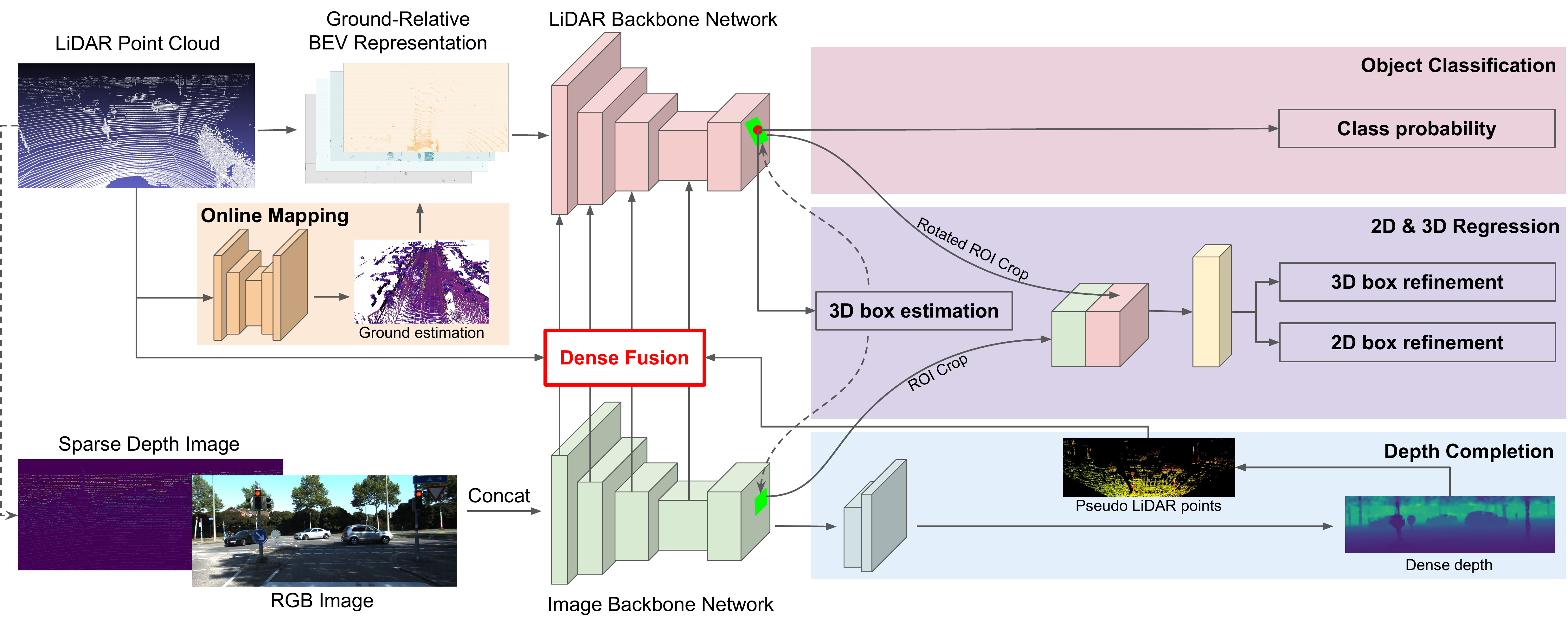}
\end{center}
   \caption{The architecture of the proposed multi-task multi-sensor fusion model for 2D and 3D object detection. Dashed arrows denote projection, while solid arrows denote data flow. Our model is a simplified two-stage detector with densely fused two-stream multi-sensor backbone networks. The first stage is a single-shot detector that outputs a small number of high-quality 3D detections. The second stage applies ROI feature fusion for more precise 2D and 3D box regression. Ground estimation is explored to incorporate geometric ground prior to the LiDAR point cloud. Depth completion is exploited to learn better cross-modality feature representation and achieve dense feature map fusion by transforming predicted dense depth image into dense pseudo LiDAR points. The whole model can be learned end-to-end.}
\label{fig:model}
\end{figure*}

\section{Multi-Task Multi-Sensor Detector}
One of the fundamental tasks in autonomous driving is to be able to perceive the scene in real-time.  
In this paper we propose a multi-task multi-sensor fusion model for the task of 3D object detection. We refer the reader to Fig. \ref{fig:model} for an illustration of the overall architecture. Our model has the following highlights. 
First, we design a multi-sensor architecture that combines point-wise and ROI-wise feature fusion. 
Second, our integrated ground estimation module reasons about the geometry of the scene. 
Third, we exploit the task of depth completion to learn better multi-sensor features and achieve dense point-wise feature fusion. 
As a result, the whole model can be learned end-to-end by exploiting a multi-task loss. Importantly, it achieves superior detection accuracy over the state of the art, with real-time efficiency.

In the following, we first introduce the single-task fully fused multi-sensor detector architecture with point-wise and ROI-wise feature fusion. We then show how we exploit the other two auxiliary tasks to further improve 3D detection. Finally we provide details of how to train our model end-to-end.

\subsection{Fully Fused Multi-Sensor Detector}
Our  multi-sensor detector takes a LiDAR point cloud and an RGB image as input. It then applies a two-stream architecture as the backbone network with point-wise feature fusion  at multiple layers. After the backbone network, the detector directly outputs high-quality 3D object detections via convolution thanks to multi-scale feature fusion. We then perform ROI-wise feature fusion via precise ROI feature extraction, and feed the fused ROI feature to a refinement module to produce very accurate 2D and 3D detections. Since the high-quality 3D detections are predicted via a fully convolutional network,  the refinement network with ROI feature fusion only has to process a small number of detections (typical fewer than $20$ on KITTI). This makes our two stage architecture very efficient.

\paragraph{Input representation:}
We use the voxel based LiDAR representation of \cite{contfuse} due to its efficiency. In particular, we voxelize the point cloud into a 3D occupancy grid, where the voxel feature is computed via 8-point linear interpolation on each LiDAR point. This LiDAR representation has the advantage of capturing fine-grained point density information efficiently. We consider the resulting 3D volume as Bird's-Eye-View (BEV) representation by treating the height slices as feature channels. This allow us to reason   in  2D BEV space. This simplification brings significant efficiency gains with no performance drop.
We simply use the RGB  image as input for the camera stream. When we exploit the auxiliary task of depth completion, we additionaly add a sparse depth image generated by projecting the LiDAR to the image plane.

\paragraph{Network architecture:}
The backbone network follows a typical two-stream architecture to process multi-sensor data. We use a 2D fully  convolutional residual network \cite{resnet} as feature extractor. Specifically, for the image stream we use a ResNet-18 \cite{resnet} architecture until the fourth residual block. Each block contains 2 residual layers with number of feature maps increasing from 64 to 512 linearly. For the LiDAR stream, we use a customized residual network which is deeper and thinner than ResNet-18 for a better trade-off between speed and accuracy. In particular, we have four residual blocks with 2, 4, 6, 6 residual layers in each, and the numbers of feature maps are 64, 128, 192 and 256. We also remove the max pooling layer before the first residual block to maintain more details in the point cloud feature. On the LiDAR stream we apply a Feature Pyramid Network (FPN) \cite{fpn} with $1\times1$ convolution and bilinear up-sampling to combine multi-scale features. Similarly we apply another FPN on the image stream  to combine multi-scale image features. 
As a result, the final feature maps on the two streams have a down-sampling factor of 4 compared with the input. On top of the feature map output from the LiDAR stream, we simply add a $1\times1$ convolution to output the object classification and 3D box regression for 3D detections. After score thresholding and oriented Non-Maximum-Suppression (NMS), a small number of high-quality 3D detections are projected to both LiDAR BEV space and 2D image space, and their ROI features are cropped from each stream's backbone feature map via precise ROI feature extraction. The two-stream ROI features are fused together and fed into a refinement module with two 256-dimension Fully Connected (FC) layers to predict the 2D and 3D box refinements for each 3D detection.

\paragraph{Point-wise Feature Fusion:}
\begin{figure}[t]
\begin{center}
\includegraphics[width=1.0\linewidth]{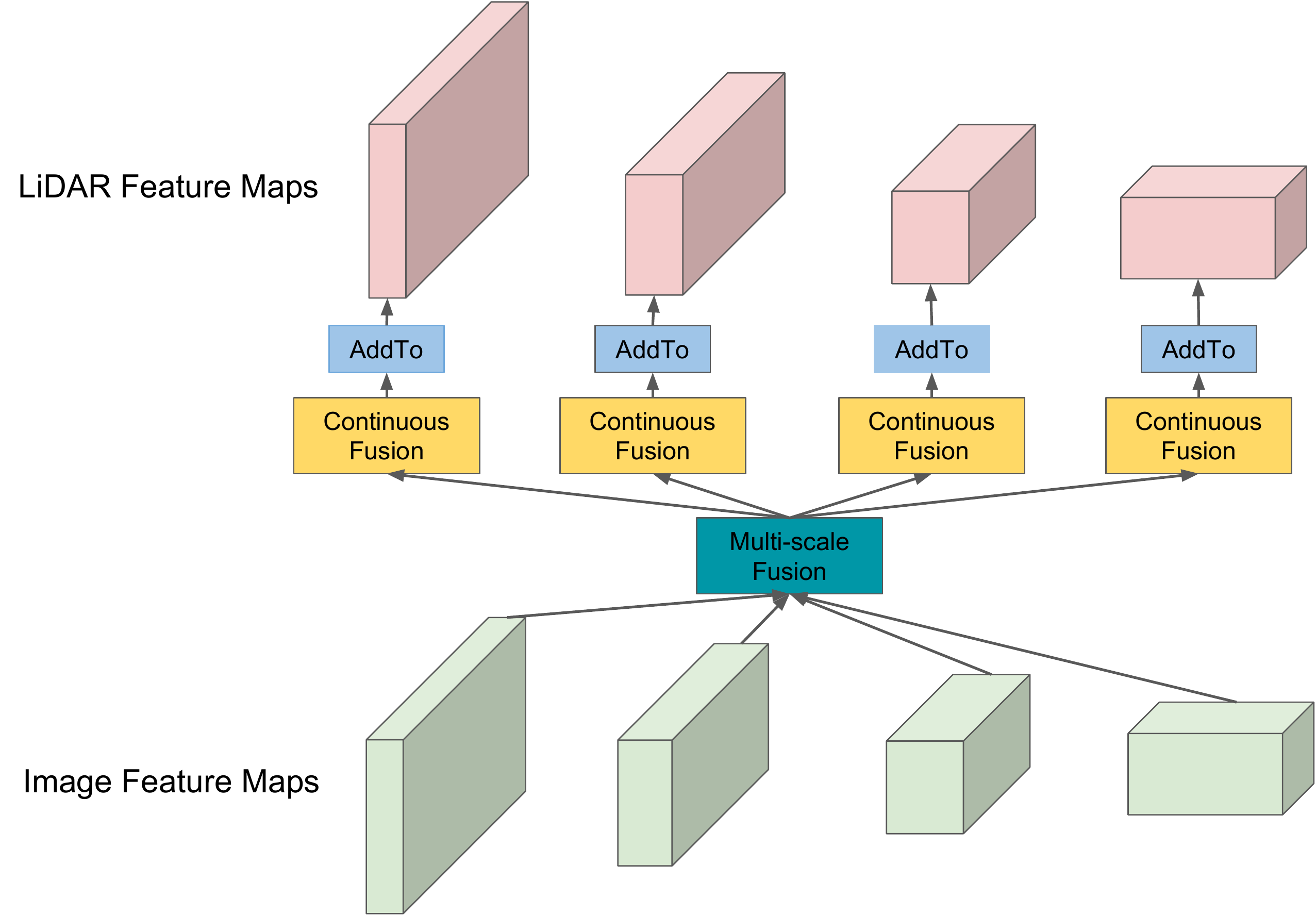}
\end{center}
   \caption{Point-wise feature fusion between multi-scale feature maps from LiDAR and image backbone networks.}
\label{fig:point}
\end{figure}

We apply point-wise feature fusion between the convolutional feature maps of LiDAR and image streams. The fusion is directed from image steam to LiDAR steam to augment BEV features with information richness of image features. We gather multi-scale features from all four blocks in the image backbone network by upsampling the low resolution maps and element-wisely add them together. These multi-scale image features are then fused to each block of the LiDAR backbone network. Fig. \ref{fig:point} shows an example depicting fusion of multi-scale image features to the first block of LiDAR backbone network.

To fuse multi-sensor convolutional feature maps, we need to find the pixel-wise correspondence between the two sensors. Inspired by \cite{contfuse}, we use continuous fusion to establish  dense and accurate correspondences between the image and BEV feature maps. For each pixel in the BEV feature map, we find its nearest LiDAR point and project the point onto the image feature map to retrieve the corresponding image feature.  We  compute the distance between the BEV pixel and LiDAR point as the geometric feature. Both image feature and geometric feature are pass as input into a Multi-Layer Perceptron (MLP) and the output is fused to BEV feature maps by element-wise addition.

\paragraph{ROI-wise Feature Fusion:}
\begin{figure}[t]
\begin{center}
 \includegraphics[width=1.0\linewidth]{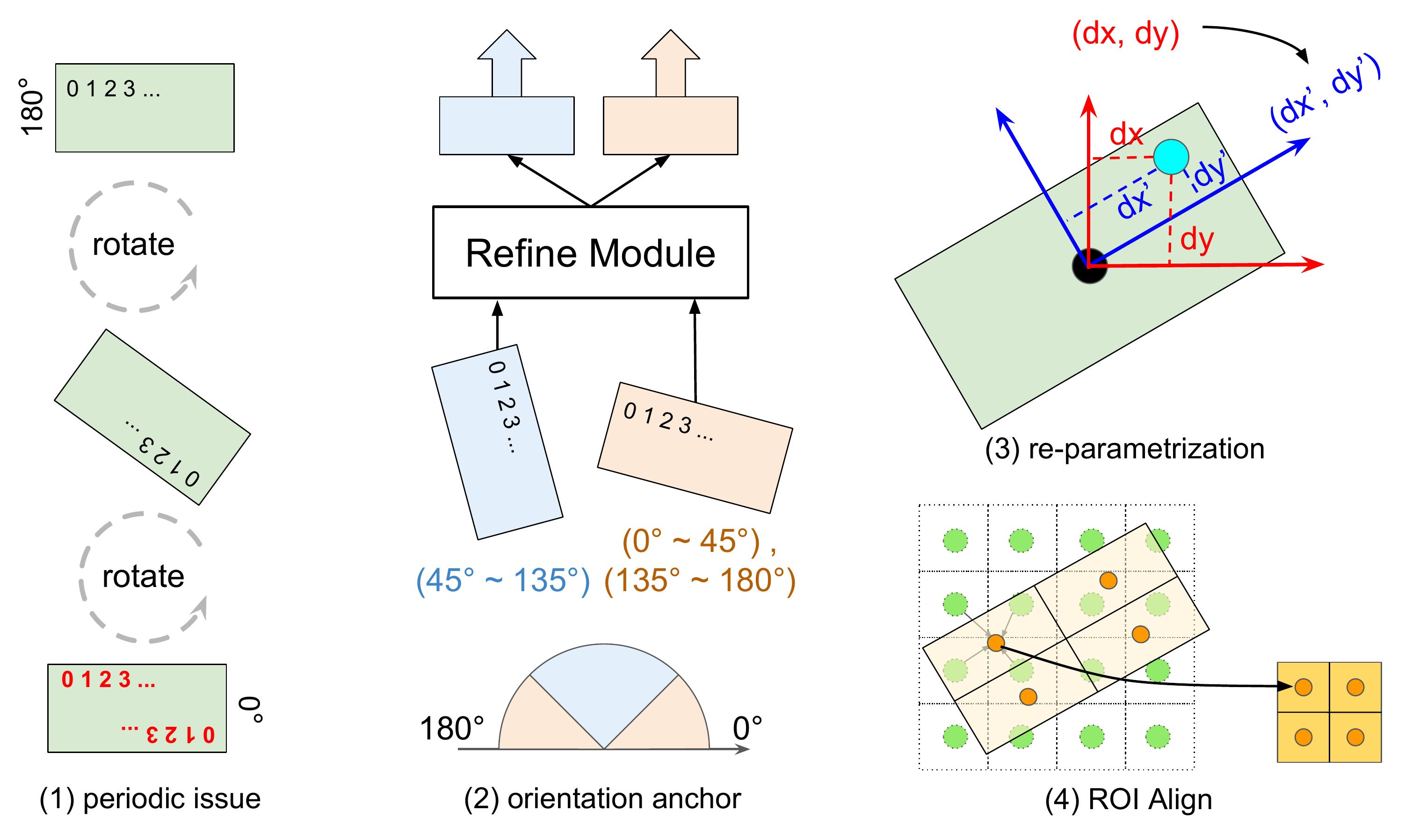}
\end{center}
   \caption{Precise rotated ROI feature extraction that takes orientation cycle into account. \textbf{(1)} The rotational periodicity causes abrupt change of order in feature extraction. \textbf{(2)} ROI refine module with two orientation anchors. An ROI is assigned to 0$^\circ$ or 90$^\circ$. They share most refining layers except for the output. \textbf{(3)} The regression target of relative offsets are re-parametrized with respect to the object orientation axes. \textbf{(4)} A $n\times n$ sized feature is extracted using bilinear interpolation (we show an example with $n=2$).}
\label{fig:roi}
\end{figure}
The motivation of the ROI-wise feature fusion is to further refine the localization precision of the high-quality 3D detections. Towards this goal, the ROI feature extraction needs to be precise so as to properly predict the relative box refinement. By projecting a 3D detection onto the image and BEV feature maps, we get an axis-aligned image ROI and an oriented BEV ROI. Feature extraction on axis-aligned image ROI is straight-forward. However, there are two new issues arising from oriented BEV ROI (see Fig. \ref{fig:roi}). First, the periodicity of the ROI orientation causes the abrupt change of feature extraction order at the cycle boundary. To solve this issue, we propose an oriented ROI feature extraction module with anchors. Given an oriented ROI, we first assign it to one of the two orientation anchors, 0 or 90 degrees. All ROIs belonging to an anchor have a consistent feature extraction order. The two anchors share the refinement net except for the output layer. Second, when the ROI is rotated, its location offsets have to be represented in the rotated coordinates as well. To implement this, we first compute the location offset in the original coordinates, and then rotate them to be aligned with the ROI. Similar to ROIAlign\cite{maskrcnn}, we extract bilinearly interpolated feature from a $n\times n$ regular grid in the ROI (in practice we use $n=5$).

\subsection{Multi-Task Learning for 3D Detection}

In this paper we exploit two auxiliary tasks to improve 3D object detection, namely ground estimation and depth completion. They help in different ways: ground estimation provides geometric  priors to enhance the LiDAR point clouds. Depth completion guides the image network to learn better cross-modality feature representations. Furthermore, it provides dense point-wise feature fusion.

\subsubsection{Ground estimation}
Mapping is an important task for autonomous driving, and in most cases the map building process is done offline. However, online mapping is appealing for that it decreases the system's dependency on offline built maps and increases the system's robustness. Here we focus on one basic sub-task in mapping of ground estimation, which is to estimate the road geometry on-the-fly from a single LiDAR sweep. We formulate the task as a regression problem, where we estimate the ground height value for each voxel in the BEV space. This formulation is more accurate than plane based parametrization \cite{3dop, birdnet}, as in practice the road is often   curved especially when we look far ahead.

\paragraph{Network architecture:}
We apply a small U-shaped Fully Convolutional Network (FCN) to estimate the normalized voxel-wise ground geometry at an inference time of 8 ms. We chose a U-Net architecture \cite{unet} since it outputs prediction at the same resolution as the input, and is good at maintaining low-level details.

\paragraph{Map fusion:}
Given a voxel-wise ground estimation, we first extract point-wise ground height by looking for the point index during voxelization. We then subtract it from each LiDAR point's $Z$ axis value and generate a new LiDAR BEV representation (relative to ground), which is fed to the LiDAR backbone network. On the first stage regression output, we add the ground height back to the predicted $Z$ term. The on-the-fly predicted ground geometry helps make 3D object localization easier because traffic participants, which are our objects of interest, all  lay on the ground.

\subsubsection{Depth completion}
LiDAR provides long range 3D information for accurate 3D object detection. However, the observation is  sparse especially at long range. Here, we propose to densify LiDAR observations by depth completion by exploiting both LiDAR and  images. Specifically, 
given the projected (into the image plane) depth observation from the LiDAR and a camera image, the model outputs
dense depth at the same resolution as the input image.

\paragraph{Sparse depth image from LiDAR projection:}
We first generate a three-channel sparse depth image from the LiDAR data, representing the sub-pixel offsets and the depth value. Specifically, we  project each LiDAR point $(x, y, z)$ to the camera space, denoted as $(x_{cam}, y_{cam}, z_{cam})$ (the $Z$ axis points to the front of the camera), where $z_{cam}$ is the depth of the LiDAR point in camera space. We then project the point from camera space to image space, denoted as $(x_{im}, y_{im})$. We find the pixel $(u, v)$ closest to $(x_{im}, y_{im})$, and compute $(x_{im} - u, y_{im} - v, z_{cam}/10)$ as the value of pixel $(u, v)$ on the sparse depth image \footnote{We divide the depth value by $10$ for normalization purpose.}. 
For pixel locations with no LiDAR point, we set the pixel value to zero. After generating the sparse depth image, we concatenate it with the RGB image along the channel dimension and feed to the image backbone network.

\paragraph{Network architecture:}
The depth completion network shares the same backbone as the image backbone network, and applies four convolutional layers accompanied with two bilinear up-sampling layers to regress the dense pixel-wise depth at the same resolution with the input image.

\paragraph{Dense depth for dense point-wise feature fusion:}
As mentioned above, the point-wise feature fusion relies on LiDAR points to find the feature map correspondence. However, since LiDAR measurements are sparse by nature, the point-wise feature fusion can be sparse, especially when the image has a larger resolution than LiDAR (for example, images captured by a camera with long-focus lens). In contrast, the depth completion task provides dense depth information per image pixel, and therefore can be used as ``pseudo'' LiDAR points to find dense feature map correspondences between the two modalities. In practice, we use the dense depth prediction for point-wise fusion only on pixels where there's no true LiDAR point found.

\subsection{Joint Training}

\begin{table*}[t]
\begin{center}
\begin{tabular}{l|cc|r||ccc|ccc|ccc}
\hline
\multirow{2}{*}{Detector}& \multicolumn{2}{|c|}{Input Data}& Time & \multicolumn{3}{|c|}{2D AP (\%)} & \multicolumn{3}{|c|}{3D AP (\%)} & \multicolumn{3}{|c}{BEV AP (\%)}\\
\cline{2-13}
 & LiDAR & IMG & (ms) & easy & mod. & hard & easy & mod. & hard & easy & mod. & hard \\
\hline
SHJU-HW~\cite{sjtu1, sjtu2} & &\checkmark & 850 & 90.81 & 90.08 & 79.98 & - & - & - & - & - & - \\
RRC~\cite{rrc} & &\checkmark & 3600 & 90.61 & {\bf 90.23} & 87.44 &  - & - & - & - & - & - \\
\hline
MV3D~\cite{mv3d} & \checkmark & & 240 & 89.80 & 79.76 & 78.61 & 66.77 & 52.73 & 51.31 & 85.82 & 77.00 & 68.94 \\
VoxelNet~\cite{voxelnet} & \checkmark & & 220 &  - & - & - & 77.49 & 65.11 & 57.73 & 89.35 & 79.26 & 77.39 \\
SECOND~\cite{second} & \checkmark& & 50 & 90.40 &	88.40 &	80.21 & 83.13 &	73.66 &	66.20 & 88.07 &	79.37 &	77.95 \\
PIXOR~\cite{pixor} & \checkmark& & {\bf 35} &  - & - & - & - & - & - & 87.25 & 81.92 & 76.01 \\
PIXOR++~\cite{hdnet} & \checkmark& & {\bf 35} &  - & - & - & - & - & - & 89.38 &	83.70 & 77.97 \\
HDNET~\cite{hdnet} & \checkmark& & 50 &  - & - & - & - & - & - & 89.14 & 86.57 & 78.32 \\
\hline
MV3D~\cite{mv3d} & \checkmark&\checkmark & 360 & 90.53 & 89.17 & 80.16 & 71.09 & 62.35 & 55.12 & 86.02 & 76.90 & 68.49 \\
AVOD~\cite{avod} & \checkmark&\checkmark & 80 & 89.73 & 88.08 & 80.14 & 73.59 & 65.78 & 58.38 & 86.80 & 85.44 & 77.73 \\
ContFuse~\cite{contfuse} & \checkmark&\checkmark & 60 & - & - & - & 82.54 & 66.22 & 64.04 & 88.81 & 85.83 & 77.33 \\
F-PointNet~\cite{fpointnet} & \checkmark&\checkmark & 170 & 90.78 & 90.00 & 80.80 & 81.20 & 70.39 & 62.19 & 88.70 & 84.00 & 75.33 \\
AVOD-FPN~\cite{avod} & \checkmark&\checkmark & 100 & 89.99 & 87.44 & 80.05 & 81.94 & 71.88 & 66.38 & 88.53 & 83.79 & 77.90\\
\hline
Our MMF & \checkmark&\checkmark & 80 & {\bf 91.82} & 90.17 & {\bf 88.54} & {\bf 86.81} & {\bf 76.75} & {\bf 68.41} & {\bf 89.49} & {\bf 87.47} & {\bf 79.10} \\
\hline
\end{tabular}
\caption{Evaluation results on the testing set of KITTI 2D, 3D and BEV object detection benchmark (car). We compare with previously published detectors on the leaderboard ranked by Average Precision (AP) in the moderate setting.}
\label{tab:kitti}
\end{center}
\vspace{-5mm}
\end{table*}

We employ mutli-task loss to train our multi-sensor detector end-to-end. 

The full model outputs  object classification, 3D box estimation, 2D and 3D box refinement, ground estimation and dense depth. During training, we have detection labels and dense depth labels, while ground estimation 
is optimized indirectly by the detection loss. There are two paths of gradient transmission for ground estimation. One is from the output where ground height is added to predicted $Z$ term. The other goes through the LiDAR backbone network to the LiDAR point cloud input where ground height is subtracted from the $Z$ coordinate. 

For object classification $L_{cls}$, we use binary cross entropy on positive and negative samples. For the 3D box estimation  $L_{box}$ and 3D box refinement losses $L_{r3d}$, we parametrize a 3D object as $(x, y, z, \log(w), \log(l), \log(h), \theta)$, and apply smooth $\ell 1$ loss on each dimension for positive samples only. For 2D box refinement loss $L_{r2d}$, we parametrize a 2D object as $(x, y, \log(w), \log(h))$, and also apply smooth $\ell 1$ loss on each dimension. For dense depth prediction loss $L_{depth}$, we sum $\ell 2$ loss over all pixels. 
The total loss for training the model is then defined as follows:
\begin{align*}
Loss = L_{cls} + \lambda(L_{box} + L_{r2d} + L_{r3d}) + \gamma L_{depth}
\end{align*}
where $\lambda, \gamma$ are the weights to balance different tasks during training.

A good initialization is important to train successfully.  We therefore use the pre-trained ResNet-18 to initialize the image backbone network. For the additional channels added to the image input, we set their corresponding weights to zero.
We also pre-train the ground estimation network on TOR4D dataset \cite{pixor} with offline maps as labels and $\ell 2$ loss as objective function \cite{hdnet}.  Other networks in the model are initialized randomly. We train the model with stochastic gradient descent using Adam optimizer \cite{adam}. 

\section{Experiments}
In this section, we first evaluate the proposed method on the KITTI 2D/3D/BEV object detection benchmarks \cite{kitti}. We  also provide a detailed ablation study to analyze the gains bring by multi-sensor fusion and multi-task learning. We then evaluate on the more challenging TOR4D multi-class BEV object detection benchmark \cite{pixor}.

\subsection{Object Detection on KITTI}

\begin{table*}[t]
\small
\begin{center}
\addtolength{\tabcolsep}{-0pt}
\begin{tabular}{l|cc|ccc||ccc|ccc|ccc}
\hline
\multirow{2}{*}{Model} & \multicolumn{2}{|c|}{Multi-Sensor} & \multicolumn{3}{|c||}{Multi-Task} & \multicolumn{3}{|c|}{2D AP (\%)} & \multicolumn{3}{|c|}{3D AP (\%)} & \multicolumn{3}{|c}{BEV AP (\%)}\\
\cline{2-15}
 & pt & roi & map & dep & depf & easy & mod. & hard & easy & mod. & hard & easy & mod. & hard \\
\hline
LiDAR only & & & & & &93.44&87.55&84.32&81.50&69.25&63.55&88.83&82.98&77.26\\
\hline
+image & \checkmark & & & & &+2.95&+1.97&+2.76&+4.62&+5.21&+3.35&+0.70&+2.39&+1.25\\
+map & \checkmark & & \checkmark & & &+3.06&+2.20&+3.33&+5.24&+7.14&+4.56&+0.36&+3.77&+1.59\\
+refine & \checkmark & \checkmark & \checkmark & & &+3.94&\bf +2.71&+4.66&\bf +6.43&+8.62&+12.03&+7.00&+4.81&+2.12\\
+depth & \checkmark & \checkmark & \checkmark & \checkmark & &\bf +4.69& +2.65&+4.64&+6.34&\bf +8.64&\bf +12.06&+7.74&+5.16&+2.26\\
\hline
full model & \checkmark & \checkmark & \checkmark & \checkmark & \checkmark & +4.61&+2.67&\bf +4.68&+6.40&+8.61&+12.02&\bf +7.83&\bf +5.27&\bf +2.34\\
\hline
\end{tabular}
\caption{Ablation study on KITTI object detection benchmark (car) training set with four-fold cross validation. {\it pt}: point-wise feature fusion. {\it roi}: ROI-wise feature fusion. {\it map}: online mapping. {\it dep}: depth completion. {\it depf}: dense fusion with dense depth.}
\label{tab:kitti_ablation}
\end{center}
\vspace{-5mm}
\end{table*}

\paragraph{Dataset and metric:}
KITTI's object detection dataset has 7,481 frames for training and 7,518 frames for testing. We evaluate our approach on ``Car'' class. We  apply the same data augmentation as \cite{contfuse} during training, which utlizes random translation, orientation and scaling on LiDAR point clouds and camera images. For multi-task training, we also leverage the dense depth labels from the intersection of KITTI's depth completion  and object detection datasets. KITTI's detection metric is defined as Average Precision (AP) averaged over 11 points on the Precision-Recall (PR) curve. The evaluation criterion for cars is 0.7 Intersection-Over-Union (IoU) in 2D, 3D or BEV. KITTI also divides labels into three subsets (easy, moderate and hard) according to the object size, occlusion and truncation levels, and ranks methods by AP in the moderate setting.

\paragraph{Implementation details:}
We detect objects within 70 meters forward and 40 meters to the left and right of the ego-car, as most of the labeled objects are within this region. We voxelize the cropped point cloud into a volume of size $512\times448\times32$ as the LiDAR input representation. We also center-crop the images of different sizes into a uniform size of $370\times1224$. We train the model on a 4 GPU machine with a total batch size of 16 frames. We set the initial learning rate to 0.001 for Adam optimizer \cite{adam} and decay it after 30 and 45 epochs respectively. The training ends after 50 epochs.

\paragraph{Evaluation results:}
We compare our approach with previously published state-of-the-art detectors in Table \ref{tab:kitti}, and show that our approach outperforms competitors by a large margin in all 2D, 3D and BEV detection tasks. In 2D detection, we surpass the  best image detector RRC \cite{rrc} by 1.1\% AP in the hard setting, while being $45\times$ faster. Note that we only use a small ResNet-18 network as the image stream backbone network, which shows that 2D detection benefits a lot from exploiting the LiDAR sensor and reasoning in 3D detection. In BEV detection, we outperform the  best detector HDNET \cite{hdnet}, which also exploits ground estimation, by 0.9\% AP. The improvement mainly comes from multi-sensor fusion. In the most challenging 3D detection task (as it requires 0.7 3D IoU), we show an even larger gain over competitors. We surpass the  best detector SECOND \cite{second} by 3.09\% AP, and outperform the previously best multi-sensor detector AVOD-FPN \cite{avod} by 4.87\% AP. We believe the large gain mainly comes from the fully fused feature representation and the proposed ROI feature extraction for precise object localization.

\begin{figure}[t]
\begin{center}
\includegraphics[width=1.0\linewidth]{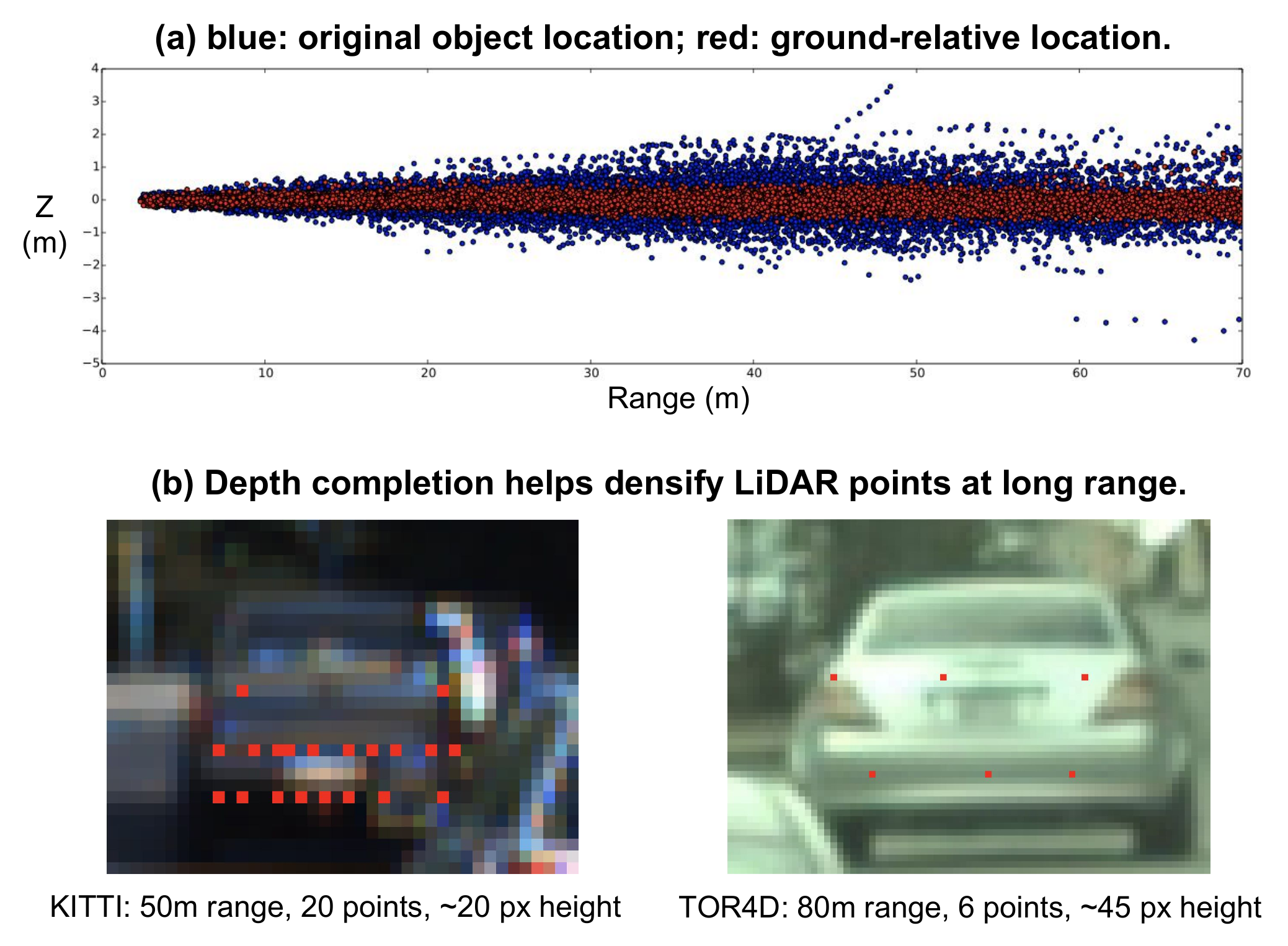}
\end{center}
   \caption{Object detection benefits from ground estimation and depth completion.}
\label{fig:mtask}
\end{figure}

\paragraph{Ablation Study:}
To analyze the effects of multi-sensor fusion and multi-task learning, we conduct an ablation study on KITTI training set. We use four-fold cross validation and accumulate the evaluation results over the whole training set. This produces stable evaluation results for apple-to-apple comparison. We show the ablation study results in Table \ref{tab:kitti_ablation}.
Our baseline model is a single-shot LiDAR only detector. Adding image stream with point-wise feature fusion brings over 5\% AP gain in 3D detection, possibly because image features provides complementary information on the $Z$ axis in addition to the BEV representation of LiDAR. 
Ground estimation improves 3D and BEV detection by 1.9\% and 1.4\% AP respectively in moderate setting. This suggests that the geometric ground prior provided by online mapping is very helpful for detection at long range (Fig. \ref{fig:mtask}), where we have very sparse 3D LiDAR measurements. 
Adding the refinement module with ROI-wise feature fusion brings consistent improvements on all three tasks, which purely comes from more precise localization. This proves the effectiveness of the proposed orientation aware ROI feature extraction. 
Lastly, the model further benefits in BEV detection from the depth completion task with better feature representations and dense fusion, which suggests that depth completion provides  complementary information in BEV space. On KITTI we do not see much gain from dense point-wise fusion using estimated depth. We hypothesize  this is   because in KITTI the captured image is at equivalent resolution of LiDAR at long range (Fig. \ref{fig:mtask}).  Therefore, there is not much juice to squeeze from another modality. However, as we will see in next section, on TOR4D benchmark where we have higher resolution camera images, we show that depth completion helps not only by multi-task learning, but also dense feature fusion.

\begin{figure*}[t]
\begin{center}
\includegraphics[width=0.33\linewidth]{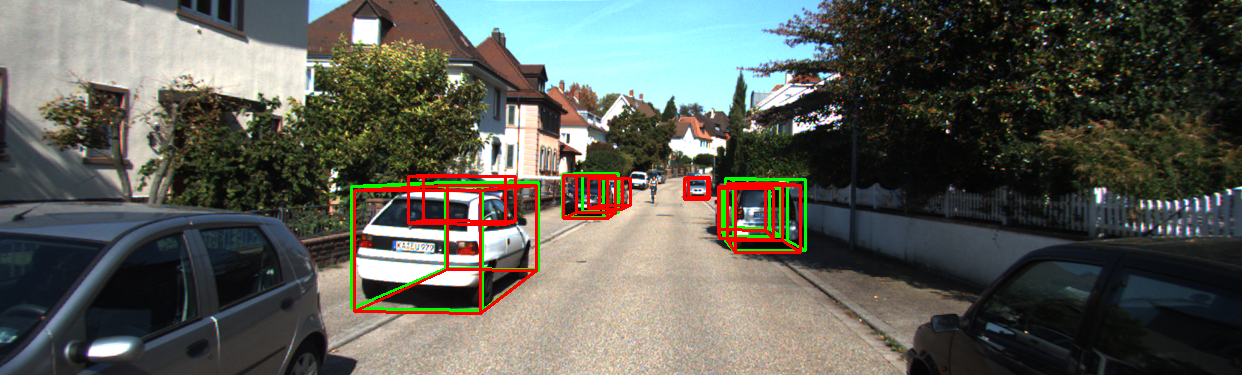} \includegraphics[width=0.33\linewidth]{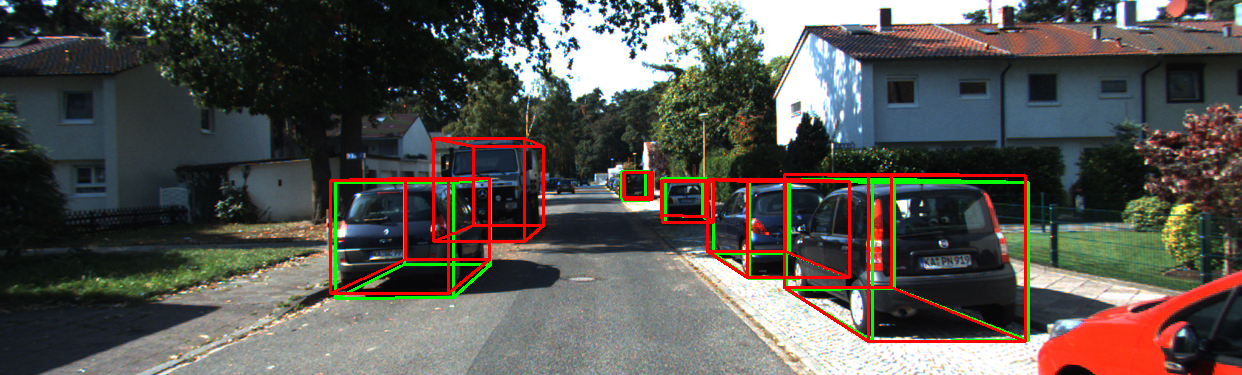} \includegraphics[width=0.33\linewidth]{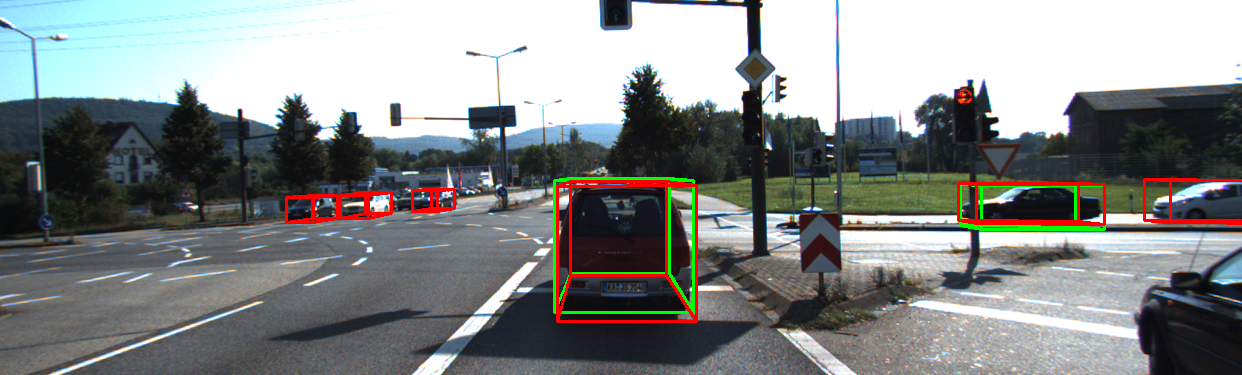}\\
\includegraphics[width=0.33\linewidth]{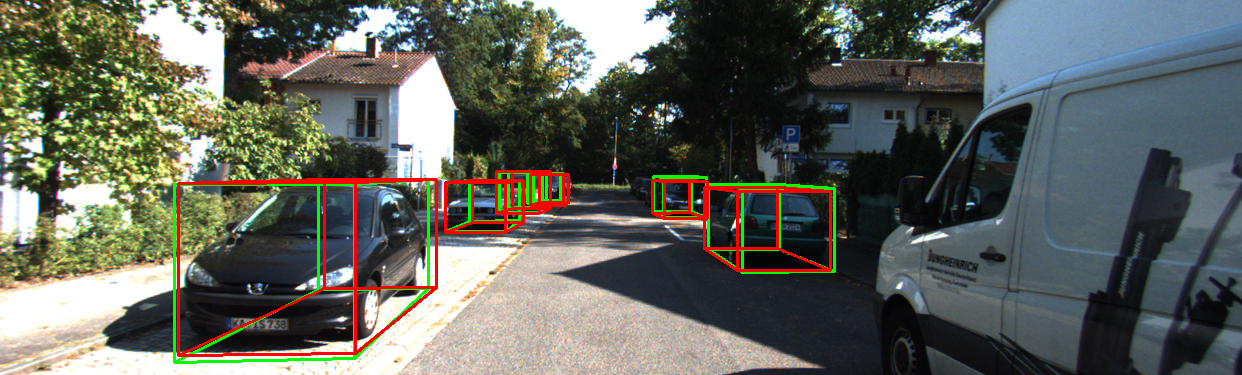} \includegraphics[width=0.33\linewidth]{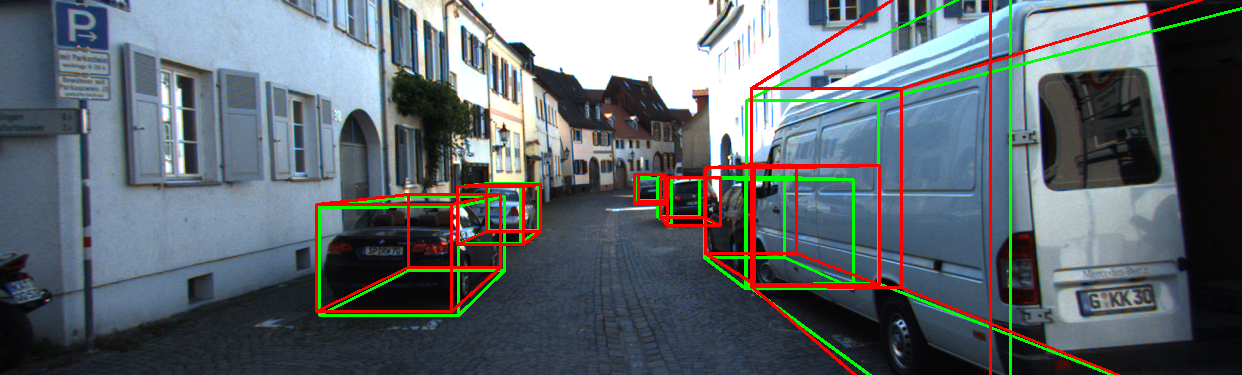} \includegraphics[width=0.33\linewidth]{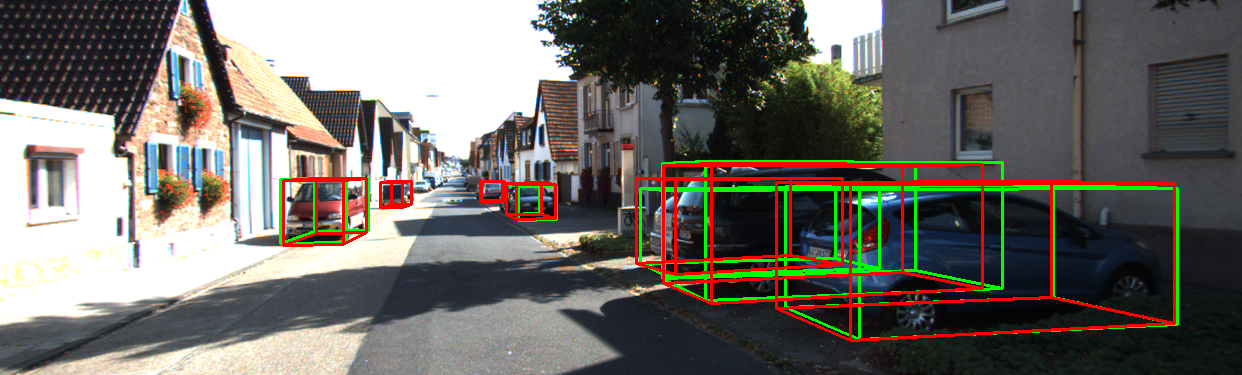}\\
\includegraphics[width=0.33\linewidth]{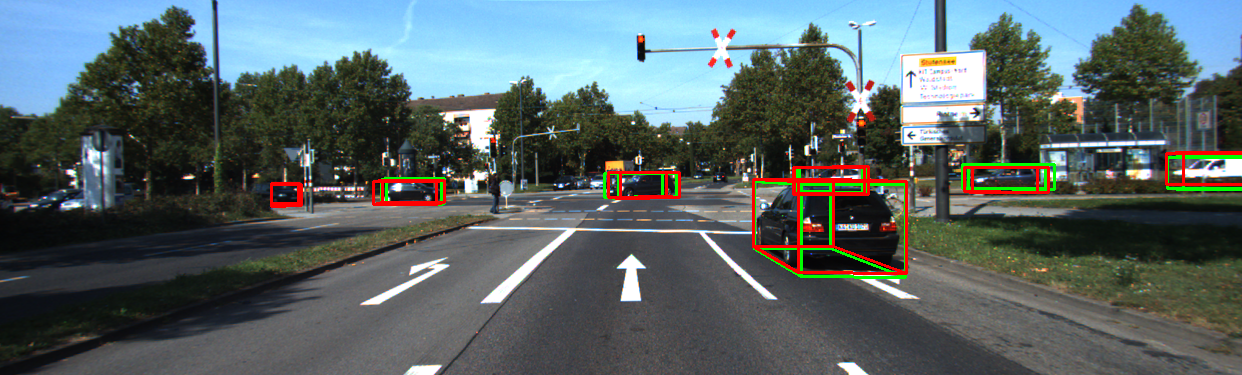} \includegraphics[width=0.33\linewidth]{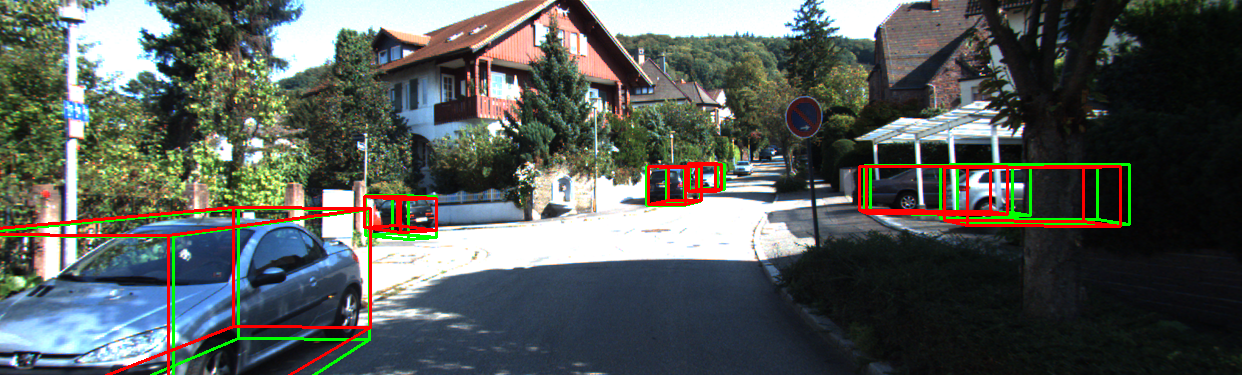} \includegraphics[width=0.33\linewidth]{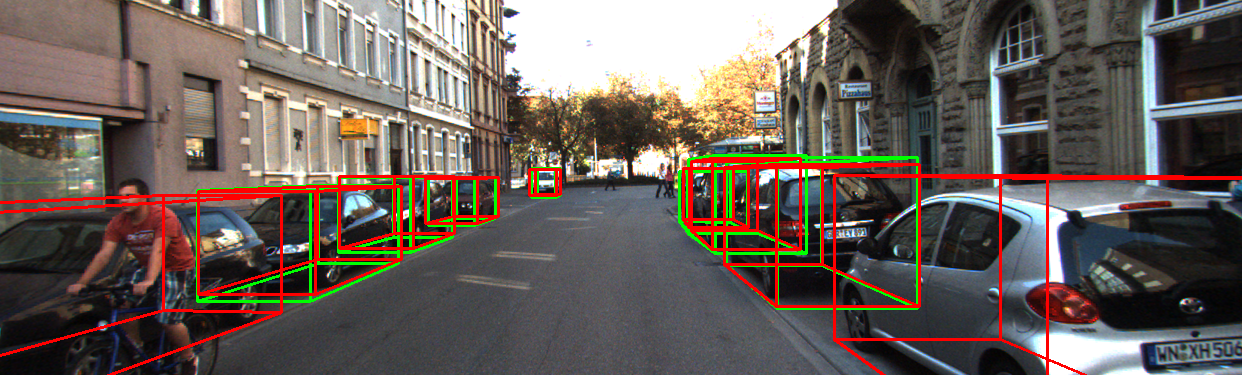}\\
\end{center}
   \caption{Qualitative results of 3D object detection (car) on KITTI benchmark. We draw object labels in green and our detections in red.}
\label{fig:demo}
\end{figure*}

\subsection{BEV Object Detection on TOR4D}


\begin{table}[t]
\small
\begin{center}
\begin{tabular}{l||c|c|c}
\hline
\multirow{2}{*}{Model} & Vehicle & Pedestrian & Bicyclist\\
\cline{2-4}
& AP$_{0.5}$ AP$_{0.7}$ & AP$_{0.3}$ AP$_{0.5}$ & AP$_{0.3}$ AP$_{0.5}$\\
\hline
ContFuse \cite{contfuse} & 95.1 \quad 83.7 & 88.9 \quad 80.7 & 72.8 \quad 58.0 \\
+dep    & 95.6 \quad 84.5 & 88.9 \quad 81.2 & 74.3 \quad 62.2 \\
+dep+depf & \bf 95.7 \quad \bf 85.4 &\bf 89.4 \quad \bf81.8&\bf76.3 \quad \bf63.1 \\
\hline
\end{tabular}
\caption{Ablation study of BEV object detection with multi-task learning on TOR4D benchmark. {\it dep}: depth completion. {\it depf}: dense fusion using estimated dense depth.}
\label{tab:tor4d_ablation}
\end{center}
\vspace{-5mm}
\end{table}

\paragraph{Dataset and metric:}
The TOR4D BEV object detection benchmark \cite{pixor} contains over 5,000 video snippets with a duration of around 20 seconds each. To generate the training and testing dataset, we  sample from different snippets at 1 Hz and 0.5Hz respectively, leading to around 100,000 training frames and around 6,000 testing frames. To validate the effectiveness of depth completion in improving object detection, we use images captured by camera with long-focus lens which provide richer information at long range (Fig. \ref{fig:mtask}). We evaluate on multi-class BEV object detection (i.e., vehicle, pedestrian and bicyclist) with a range of 100 meters distance from the ego-car. We use AP at different IoU thresholds as the metric for multi-class object detection. Specifically, we look at 0.5 and 0.7 IoU for vehicles, 0.3 and 0.5 IoU for the pedestrians and cyclists.

\paragraph{Evaluation results:}
We re-produce the previously state-of-the-art detector ContFuse \cite{contfuse} on TOR4D under our current setting. Two modifications are made to further improve the detection performance. First, we follow FAF \cite{dpt} to fuse multi-frame of LiDAR point clouds together. Second, following HDNET \cite{hdnet} we incorporate semantic and geometric High-Definition map priors to the detector. We use the new ContFuse detector as the baseline, and apply the proposed depth completion with dense fusion on top of it. As shown in Table \ref{tab:tor4d_ablation}, the depth completion task helps in two ways: multi-task learning and dense feature fusion. The former increases the bicyclist AP by an absolute 4.2\%. Since bicyclists have the fewest number of labels in the dataset, having additional multi-task supervision is particularly helpful. In terms of dense fusion with estimated depth, the performance on vehicles improves by over 5\% in terms of relative error reduction (1-AP).
The reason may be that vehicles receive more additional feature fusion compared to the other two classes (Fig. \ref{fig:mtask}).

\subsection{Qualitative Results and Discussion}
We show qualitative 3D object detection results of the proposed detector on KITTI benchmark in Fig. \ref{fig:demo}. The proposed detector is able to produce high-quality 3D detections of objects that are highly occluded or far away from the ego-car. Some of our detections are unannotated cars in KITTI.
Previous works \cite{mv3d, avod} often follow state-of-the-art 2D detection framework (like two-stage Faster RCNN \cite{faster}) to solve 3D detection. However, we argue that it may not be the optimal solution. With thousands of pre-defined anchors, the feature extraction is both slow and inaccurate. Instead we show that by detecting 3D objects in BEV space, we can produce high-quality 3D detections via a single pass of FCN (as shown in ablation study), given that we fully fuse the multi-sensor feature maps via dense fusion.

Cascade approaches \cite{fpointnet, fpccnn} suggest that 2D detection is solved better than 3D detection, and therefore use 2D detector to generate 3D proposals. However, we argue that 3D detection is actually easier than 2D. Because we detect objects in 3D metric space, we do not have to handle the problems of scale variance and occlusion reasoning that arise in 2D. Our model, using a pre-trained ResNet-18 as image network and trained from thousands of object labels, surpasses F-PointNet \cite{fpointnet}, which exploits two orders of magnitude more training data, by over 7\% AP in hard setting of KITTI 2D detection.
Multi-sensor fusion and multi-task learning are highly interleaved. In this paper we provide a way to combine them together under the same hood. In the proposed framework, multi-sensor fusion helps learn better feature representations to solve multiple tasks, while different tasks in turn provide different types of cues to make feature fusion deeper and richer.

\section{Conclusion}

We have proposed a multi-task multi-sensor detection model that jointly reasons about 2D and 3D object detection, ground estimation and depth completion. Point-wise and ROI-wise feature fusion are applied to achieve full multi-sensor fusion, while multi-task learning provides additional map prior and geometric cues enabling better representation learning and denser feature fusion. We validate the proposed method on KITTI \cite{kitti} and TOR4D \cite{pixor} benchmarks, and surpass the  state-of-the-art  in all detection tasks by a large margin. In the future, we plan to expand our multi-sensor fusion approach to exploit other sensors such as radar as well as temporal information.

{\small
\bibliographystyle{ieee}
\bibliography{egbib}

\begin{thebibliography}{10}\itemsep=-1pt

\bibitem{birdnet}
J.~Beltran, C.~Guindel, F.~M. Moreno, D.~Cruzado, F.~Garcia, and A.~de~la
  Escalera.
\newblock Birdnet: a 3d object detection framework from lidar information.
\newblock {\em IEEE International Conference on Intelligent Transportation
  Systems}, 2018.

\bibitem{mono3d}
X.~Chen, K.~Kundu, Z.~Zhang, H.~Ma, S.~Fidler, and R.~Urtasun.
\newblock Monocular 3d object detection for autonomous driving.
\newblock In {\em CVPR}, 2016.

\bibitem{3dop}
X.~Chen, K.~Kundu, Y.~Zhu, A.~G. Berneshawi, H.~Ma, S.~Fidler, and R.~Urtasun.
\newblock 3d object proposals for accurate object class detection.
\newblock In {\em NIPS}, 2015.

\bibitem{3doppami}
X.~Chen, K.~Kundu, Y.~Zhu, H.~Ma, S.~Fidler, and R.~Urtasun.
\newblock 3d object proposals using stereo imagery for accurate object class
  detection.
\newblock {\em TPAMI}, 2017.

\bibitem{mv3d}
X.~Chen, H.~Ma, J.~Wan, B.~Li, and T.~Xia.
\newblock Multi-view 3d object detection network for autonomous driving.
\newblock In {\em CVPR}, 2017.

\bibitem{fpccnn}
X.~Du, M.~H. Ang~Jr, S.~Karaman, and D.~Rus.
\newblock A general pipeline for 3d detection of vehicles.
\newblock In {\em ICRA}, 2018.

\bibitem{sjtu2}
L.~Fang, X.~Zhao, and S.~Zhang.
\newblock Small-objectness sensitive detection based on shifted single shot
  detector.
\newblock {\em Multimedia Tools and Applications}, pages 1--19, 2018.

\bibitem{kitti}
A.~Geiger, P.~Lenz, and R.~Urtasun.
\newblock Are we ready for autonomous driving? the kitti vision benchmark
  suite.
\newblock In {\em CVPR}, 2012.

\bibitem{maskrcnn}
K.~He, G.~Gkioxari, P.~Doll{\'a}r, and R.~Girshick.
\newblock Mask r-cnn.
\newblock In {\em ICCV}, 2017.

\bibitem{resnet}
K.~He, X.~Zhang, S.~Ren, and J.~Sun.
\newblock Deep residual learning for image recognition.
\newblock In {\em CVPR}, 2016.

\bibitem{adam}
D.~Kingma and J.~Ba.
\newblock Adam: A method for stochastic optimization.
\newblock In {\em ICLR}, 2015.

\bibitem{avod}
J.~Ku, M.~Mozifian, J.~Lee, A.~Harakeh, and S.~Waslander.
\newblock Joint 3d proposal generation and object detection from view
  aggregation.
\newblock In {\em IROS}, 2018.

\bibitem{contfuse}
M.~Liang, B.~Yang, S.~Wang, and R.~Urtasun.
\newblock Deep continuous fusion for multi-sensor 3d object detection.
\newblock In {\em ECCV}, 2018.

\bibitem{fpn}
T.-Y. Lin, P.~Doll{\'a}r, R.~Girshick, K.~He, B.~Hariharan, and S.~Belongie.
\newblock Feature pyramid networks for object detection.
\newblock In {\em CVPR}, 2017.

\bibitem{dpt}
W.~Luo, B.~Yang, and R.~Urtasun.
\newblock Fast and furious: Real time end-to-end 3d detection, tracking and
  motion forecasting with a single convolutional net.
\newblock In {\em CVPR}, 2018.

\bibitem{fpointnet}
C.~R. Qi, W.~Liu, C.~Wu, H.~Su, and L.~J. Guibas.
\newblock Frustum pointnets for 3d object detection from rgb-d data.
\newblock In {\em CVPR}, 2018.

\bibitem{pointnet}
C.~R. Qi, H.~Su, K.~Mo, and L.~J. Guibas.
\newblock Pointnet: Deep learning on point sets for 3d classification and
  segmentation.
\newblock In {\em CVPR}, 2017.

\bibitem{pointnet2}
C.~R. Qi, L.~Yi, H.~Su, and L.~J. Guibas.
\newblock Pointnet++: Deep hierarchical feature learning on point sets in a
  metric space.
\newblock In {\em NIPS}, 2017.

\bibitem{rrc}
J.~Ren, X.~Chen, J.~Liu, W.~Sun, J.~Pang, Q.~Yan, Y.-W. Tai, and L.~Xu.
\newblock Accurate single stage detector using recurrent rolling convolution.
\newblock In {\em CVPR}, 2017.

\bibitem{faster}
S.~Ren, K.~He, R.~Girshick, and J.~Sun.
\newblock Faster r-cnn: Towards real-time object detection with region proposal
  networks.
\newblock In {\em NIPS}, 2015.

\bibitem{rencvpr16}
Z.~Ren and E.~B. Sudderth.
\newblock Three-dimensional object detection and layout prediction using clouds
  of oriented gradients.
\newblock In {\em CVPR}, 2016.

\bibitem{rencvpr18}
Z.~Ren and E.~B. Sudderth.
\newblock 3d object detection with latent support surfaces.
\newblock In {\em CVPR}, 2018.

\bibitem{unet}
O.~Ronneberger, P.~Fischer, and T.~Brox.
\newblock U-net: Convolutional networks for biomedical image segmentation.
\newblock {\em International Conference on Medical Image Computing and
  Computer-Assisted Intervention}, 2015.

\bibitem{schwing2013box}
A.~G. Schwing, S.~Fidler, M.~Pollefeys, and R.~Urtasun.
\newblock Box in the box: Joint 3d layout and object reasoning from single
  images.
\newblock In {\em ICCV}, 2013.

\bibitem{holistic}
S.~Wang, S.~Fidler, and R.~Urtasun.
\newblock Holistic 3d scene understanding from a single geo-tagged image.
\newblock In {\em CVPR}, 2015.

\bibitem{mono3d18}
B.~Xu and Z.~Chen.
\newblock Multi-level fusion based 3d object detection from monocular images.
\newblock In {\em CVPR}, 2018.

\bibitem{second}
Y.~Yan, Y.~Mao, and B.~Li.
\newblock Second: Sparsely embedded convolutional detection.
\newblock {\em Sensors}, 18(10):3337, 2018.

\bibitem{hdnet}
B.~Yang, M.~Liang, and R.~Urtasun.
\newblock Hdnet: Exploiting hd maps for 3d object detection.
\newblock In {\em 2nd Conference on Robot Learning (CoRL)}, 2018.

\bibitem{pixor}
B.~Yang, W.~Luo, and R.~Urtasun.
\newblock Pixor: Real-time 3d object detection from point clouds.
\newblock In {\em CVPR}, 2018.

\bibitem{sjtu1}
S.~Zhang, X.~Zhao, L.~Fang, F.~Haiping, and S.~Haitao.
\newblock Led: Localization-quality estimation embedded detector.
\newblock In {\em IEEE International Conference on Image Processing}, 2018.

\bibitem{voxelnet}
Y.~Zhou and O.~Tuzel.
\newblock Voxelnet: End-to-end learning for point cloud based 3d object
  detection.
\newblock In {\em CVPR}, 2018.

\end{thebibliography}
}

\section*{Supplementary Materials}
We provide more quantitative and qualitative results on KITTI \cite{kitti} and TOR4D \cite{pixor} benchmarks.

\begin{figure*}[t]
\begin{center}
\includegraphics[width=0.33\linewidth]{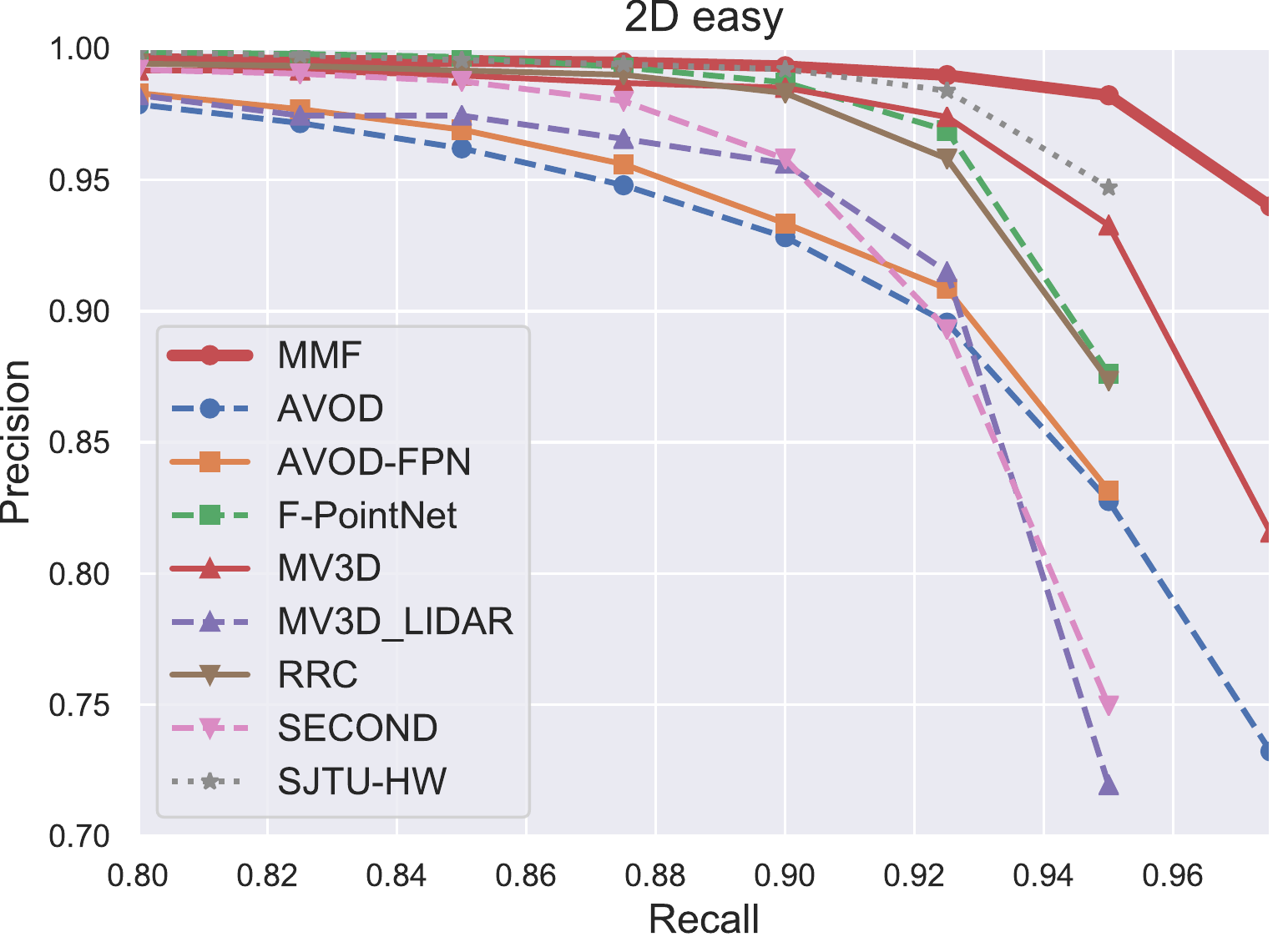} \includegraphics[width=0.33\linewidth]{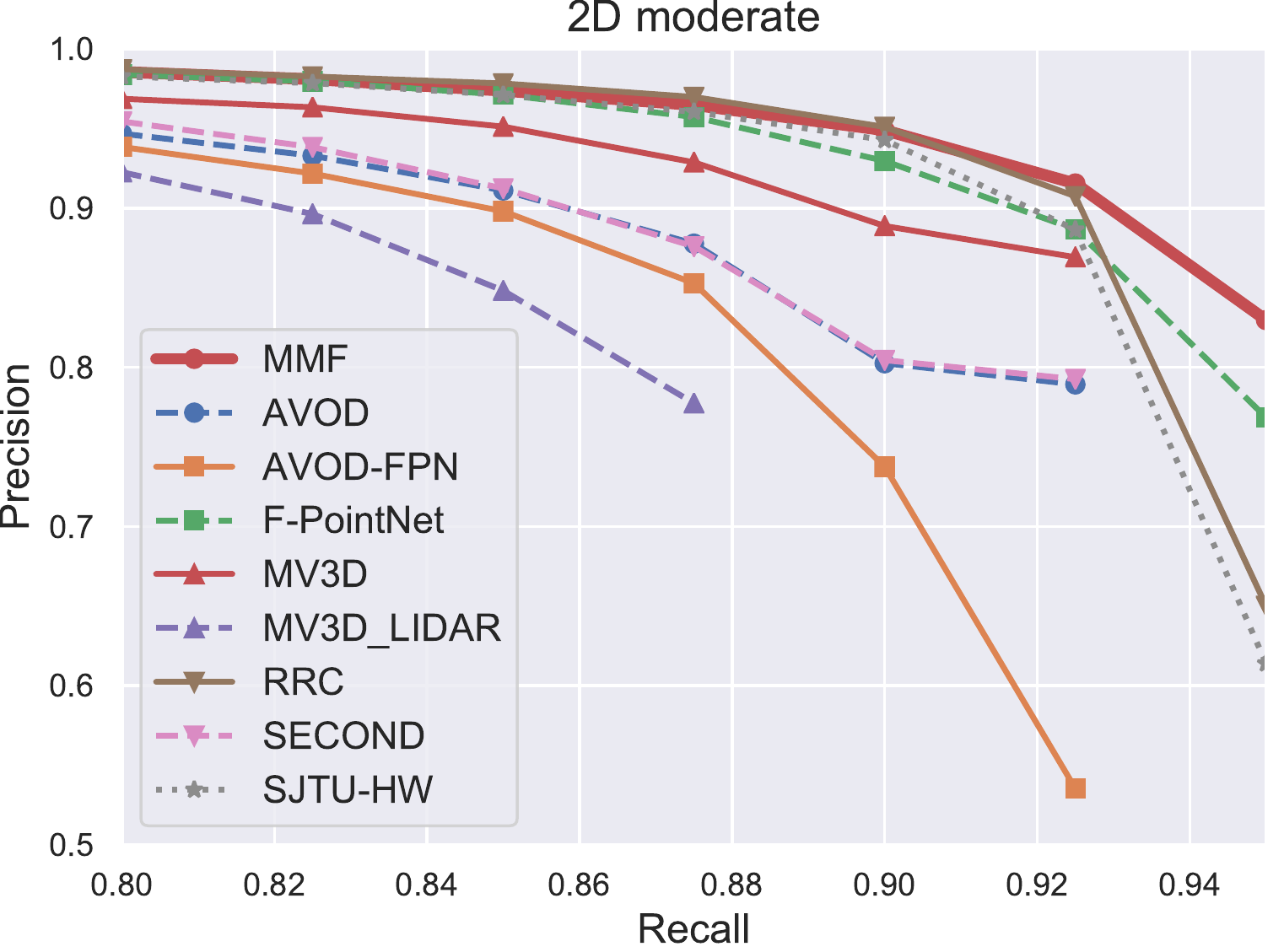} \includegraphics[width=0.33\linewidth]{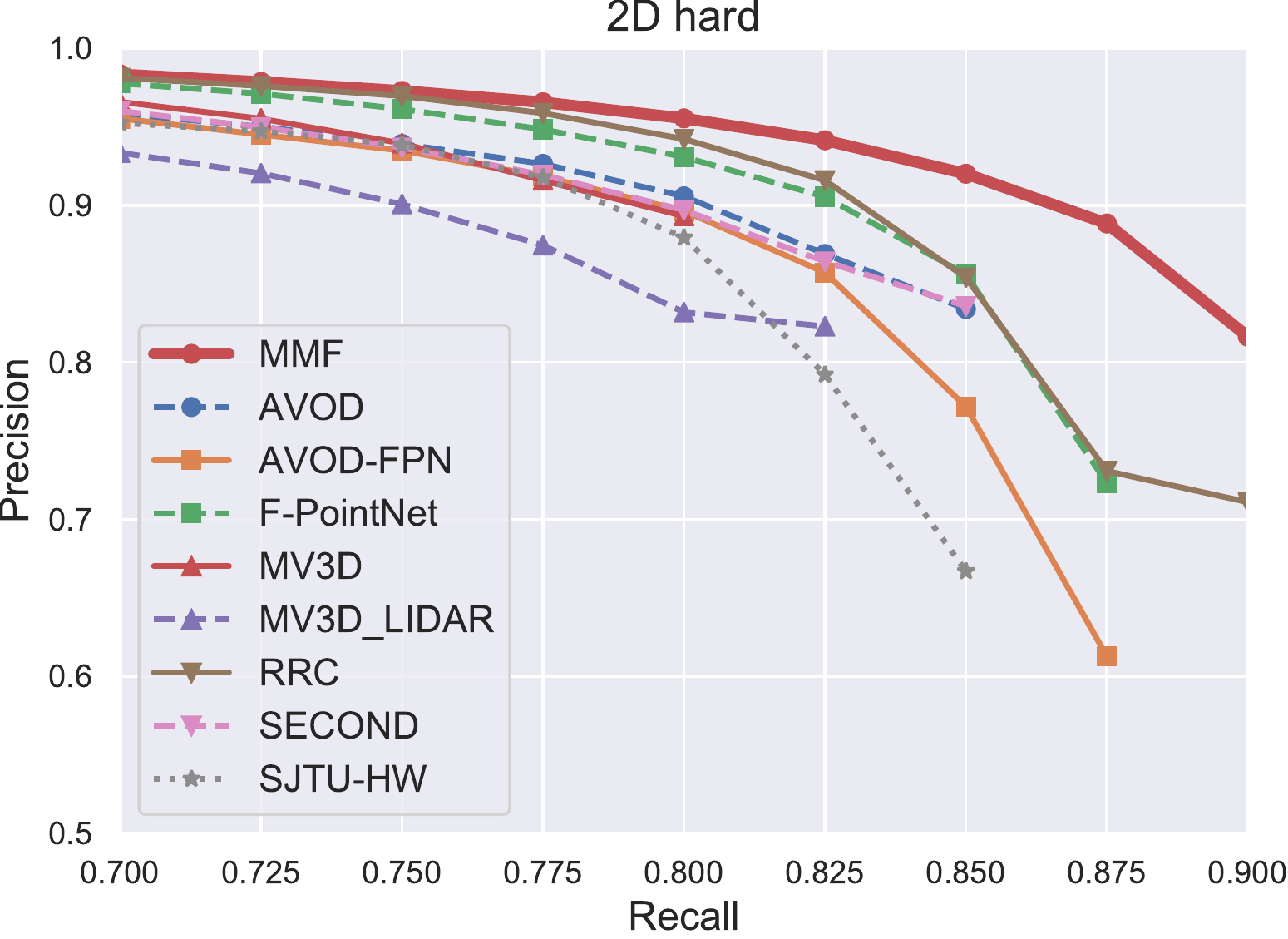}\\
\includegraphics[width=0.33\linewidth]{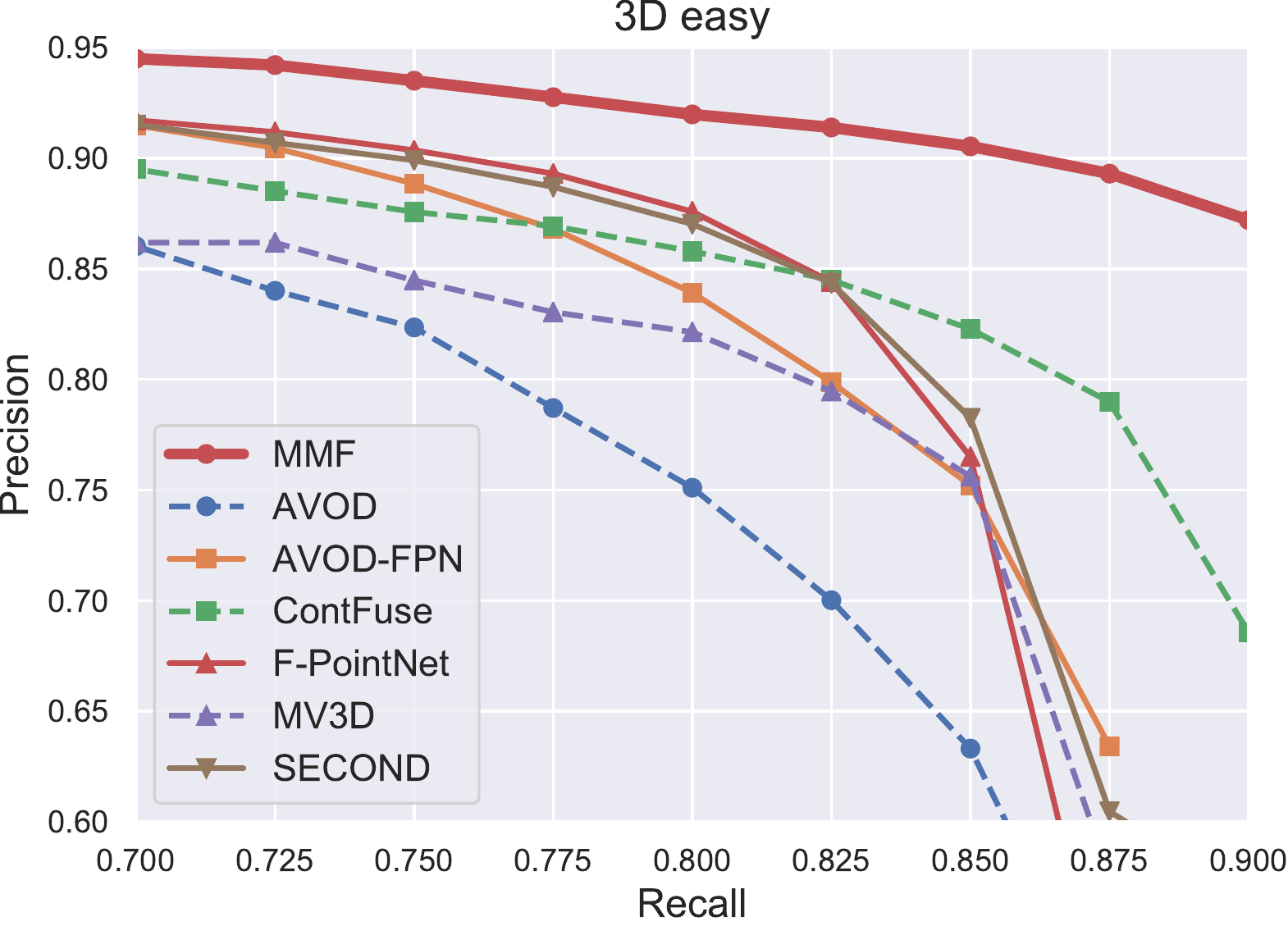} \includegraphics[width=0.33\linewidth]{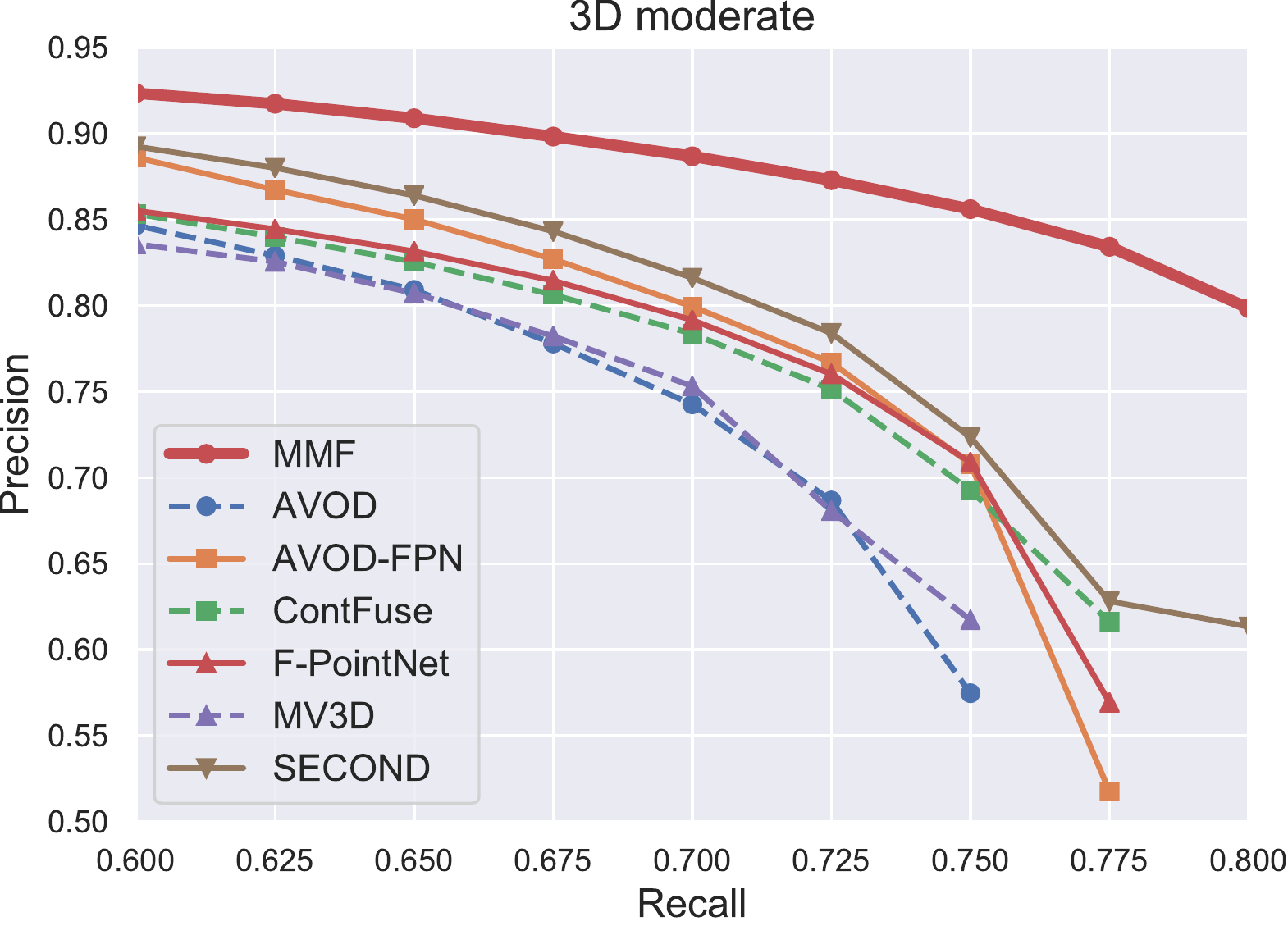} \includegraphics[width=0.33\linewidth]{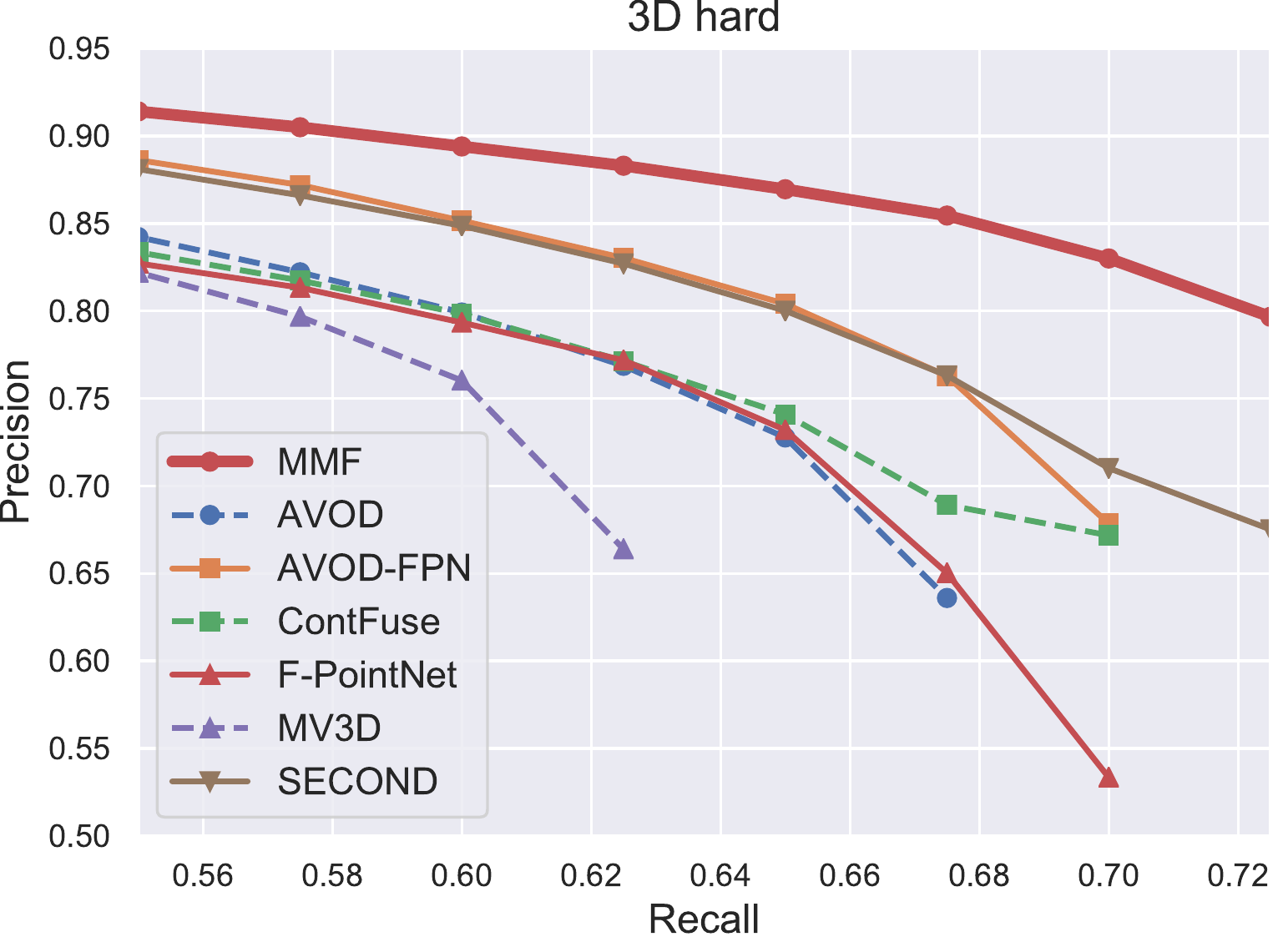}\\
\includegraphics[width=0.33\linewidth]{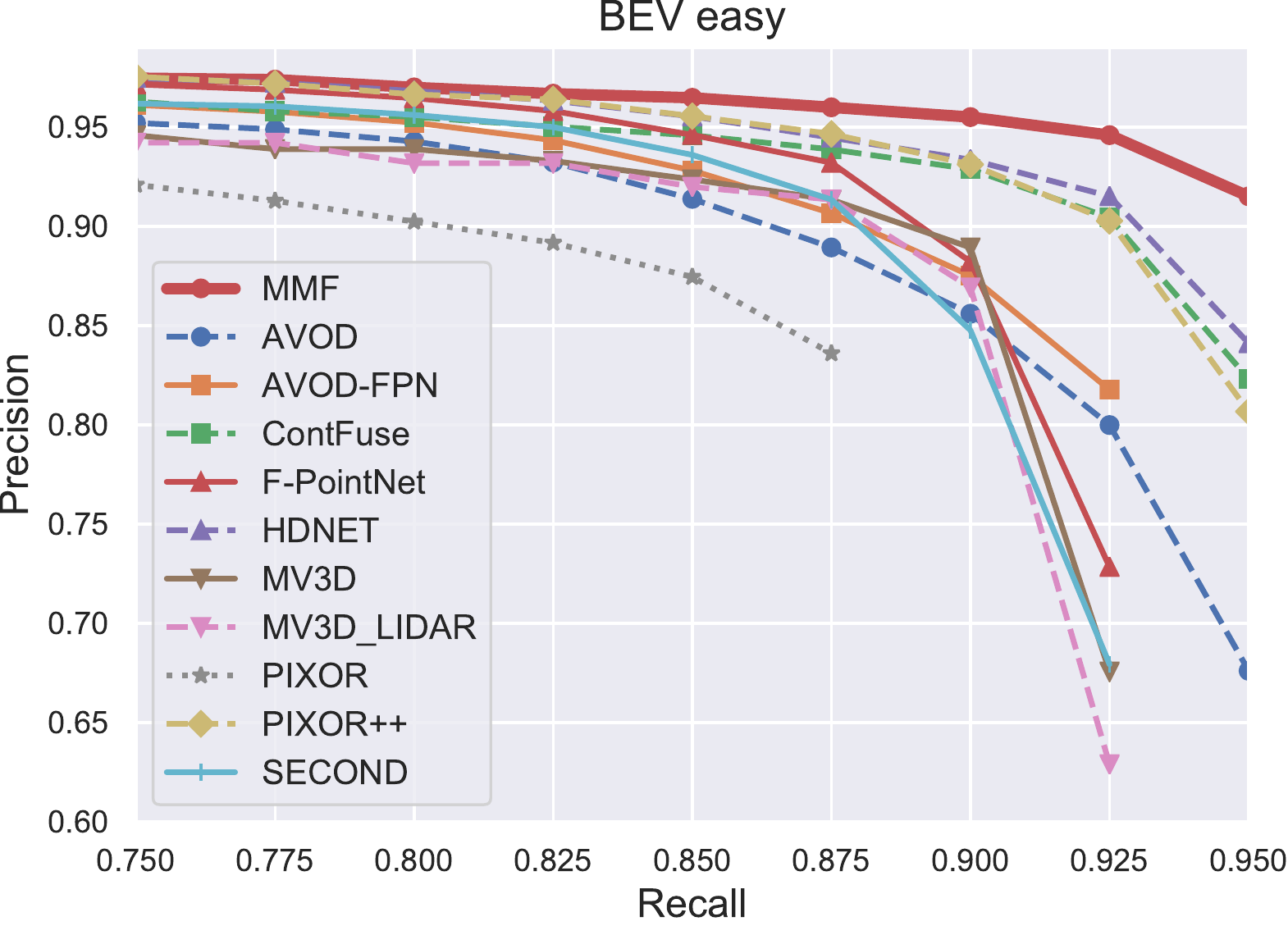} \includegraphics[width=0.33\linewidth]{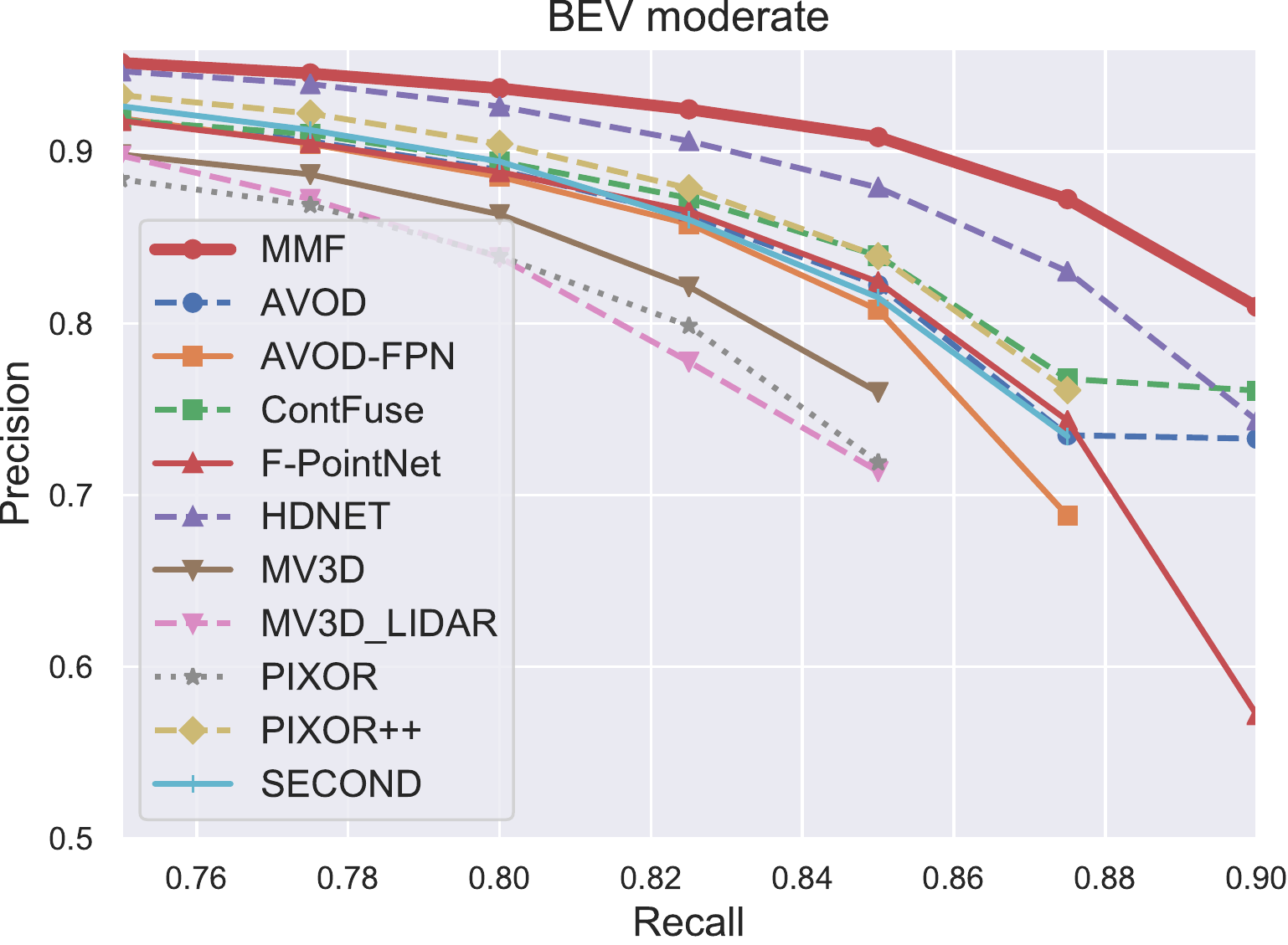} \includegraphics[width=0.33\linewidth]{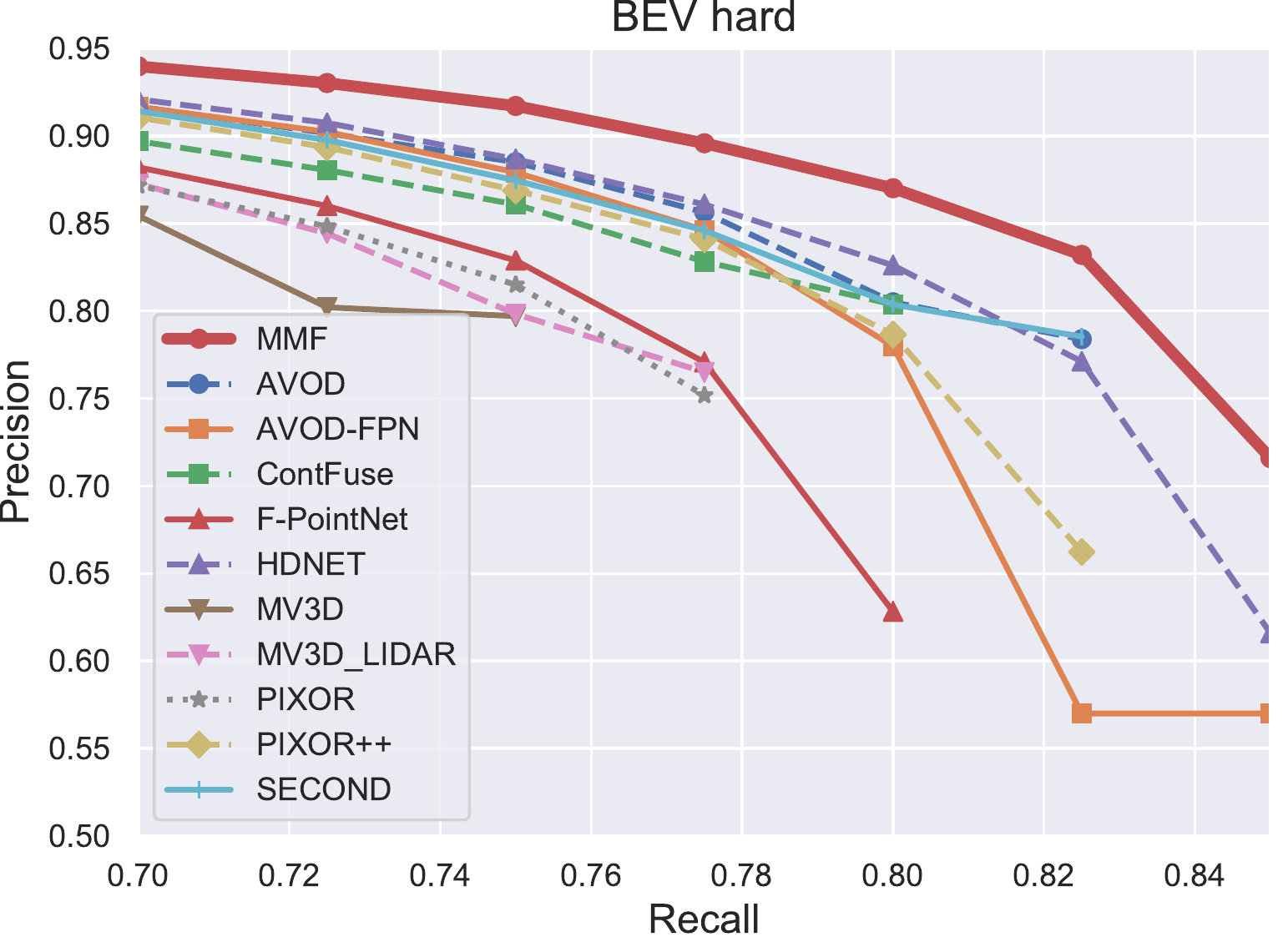}
\end{center}
   \caption{PR curve comparison between the proposed \textbf{MMF} and other state-of-the-art in 2D/3D/BEV car detection on KITTI testing set.}
\label{fig:kitti_pr}
\end{figure*}

Fig. \ref{fig:kitti_pr} shows the PR curves of the proposed detector as well as other state-of-the-art approaches in 2D/3D/BEV car detection on KITTI test set for a more comprehensive comparison. In all detection settings, the proposed detector shows consistent advantage in terms of precision rate, which proves the effectiveness of the proposed joint model in producing high-quality detections.

\begin{figure*}[t]
\begin{center}
\includegraphics[width=0.3\linewidth]{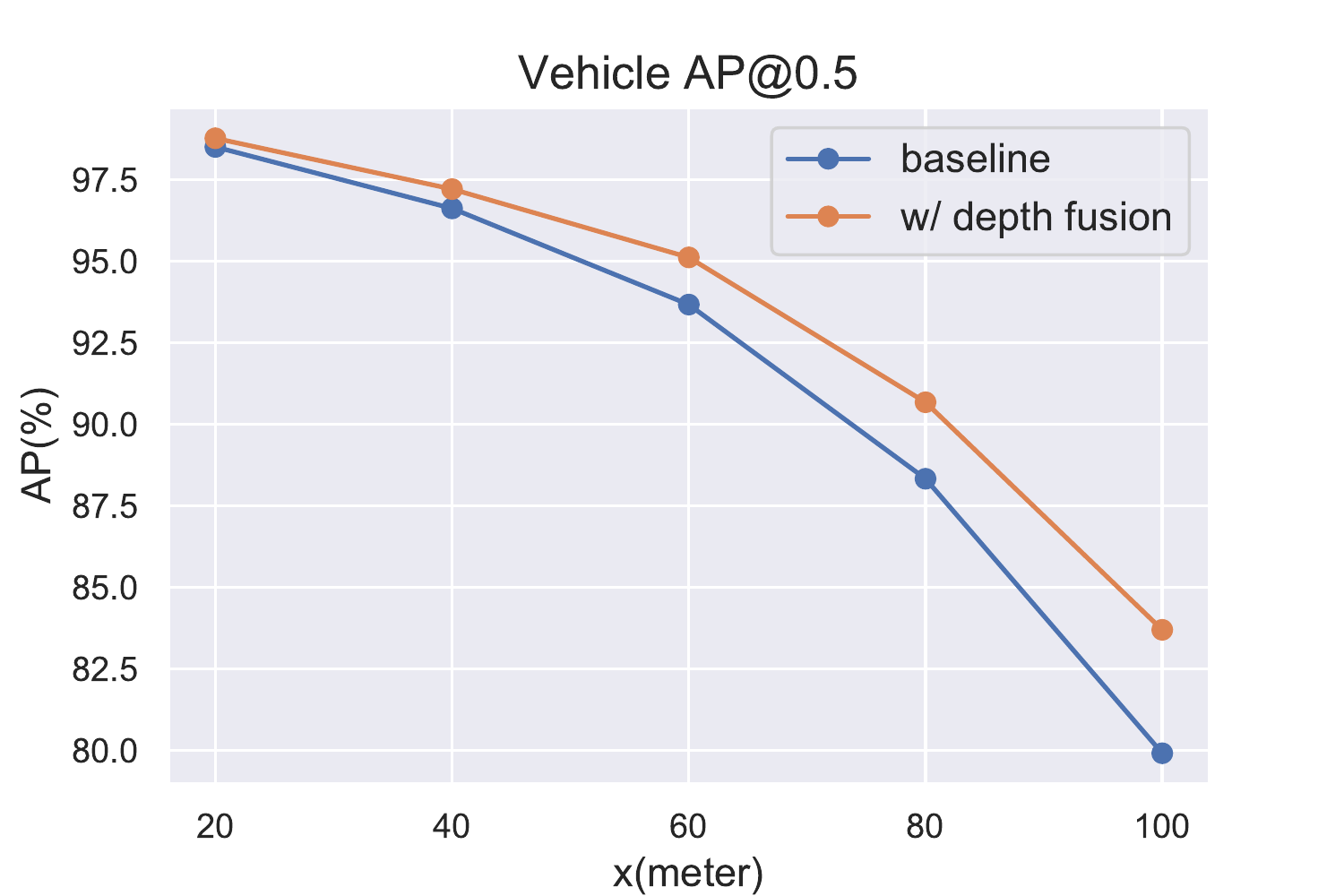} \includegraphics[width=0.3\linewidth]{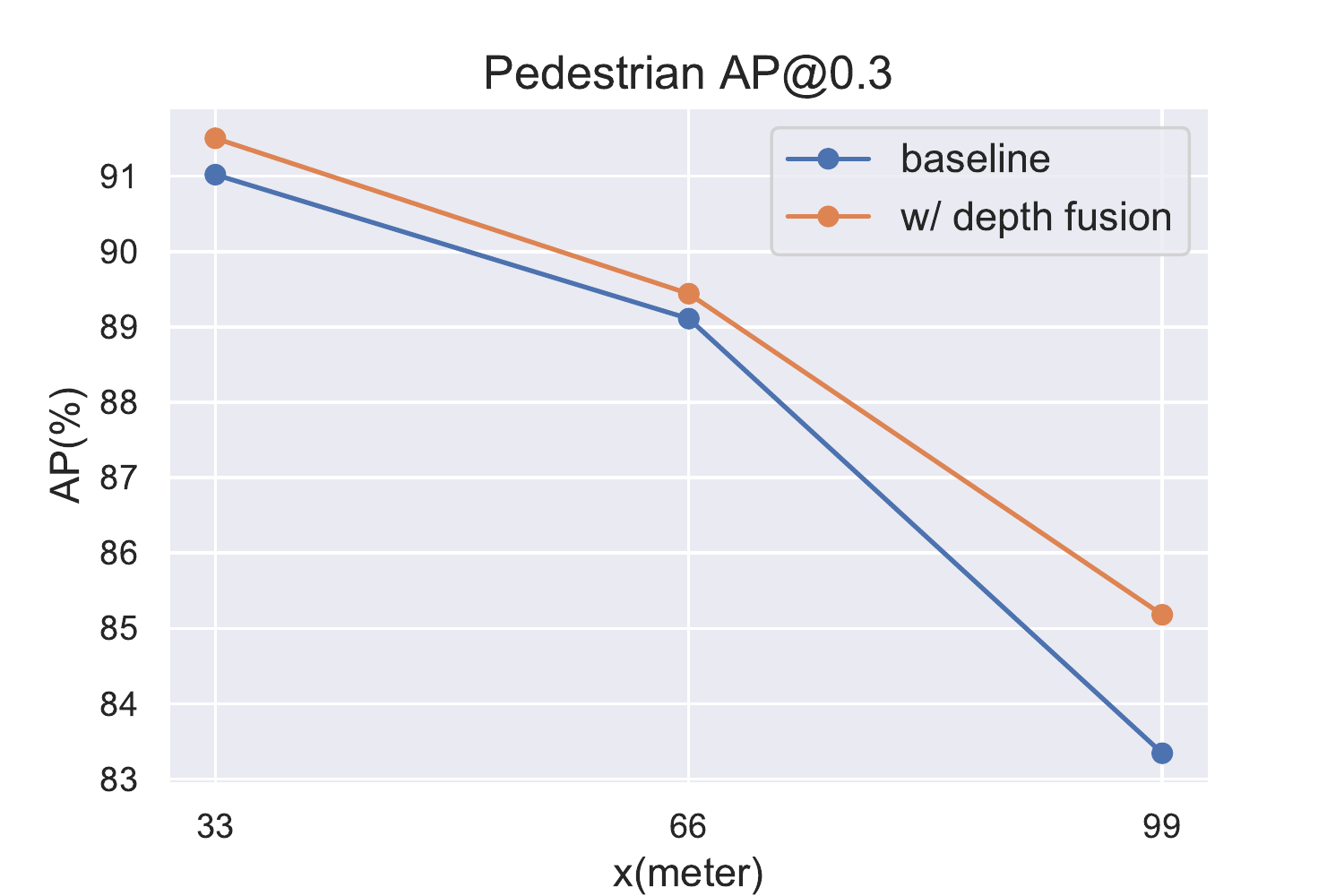} \includegraphics[width=0.3\linewidth]{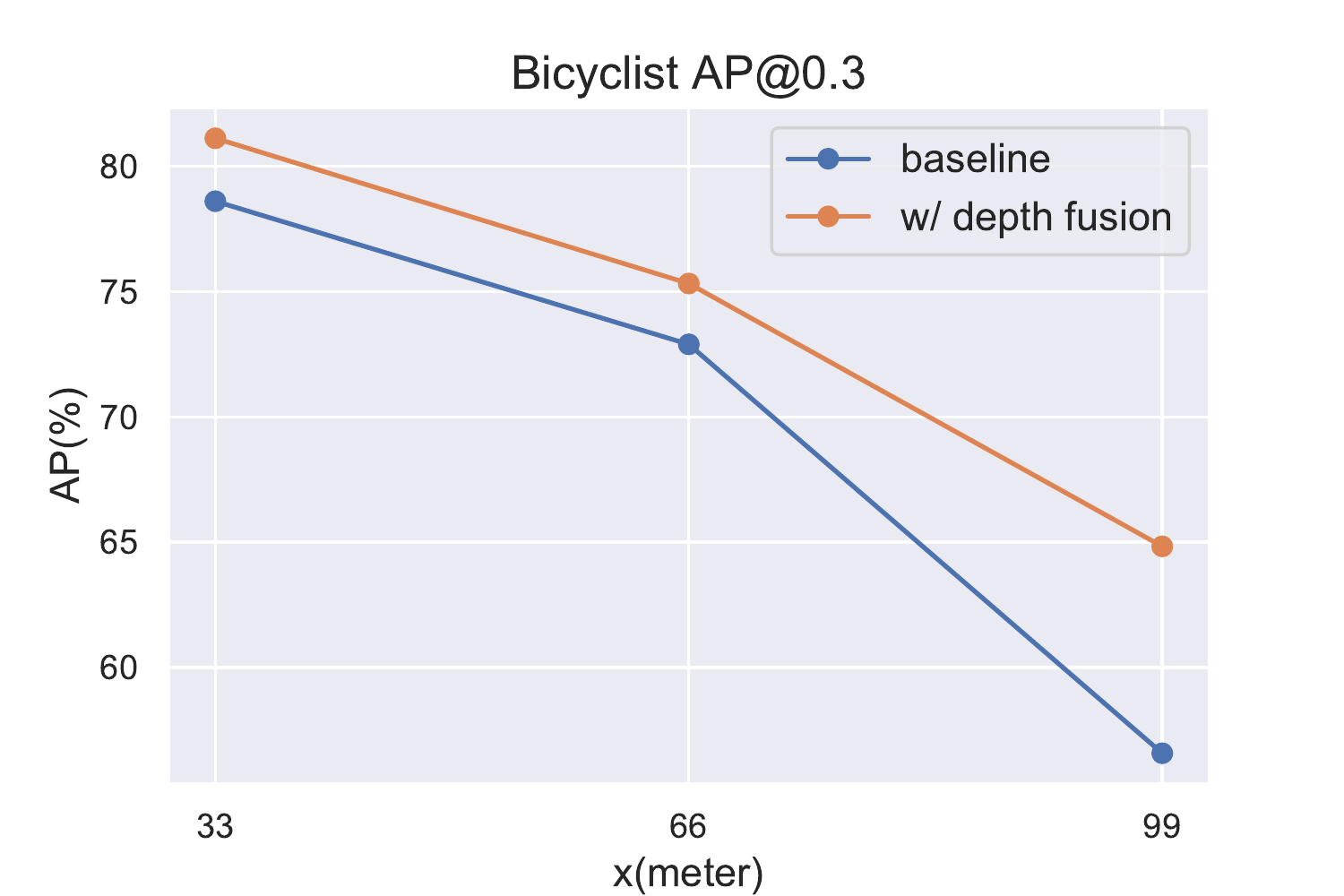}\\
\includegraphics[width=0.3\linewidth]{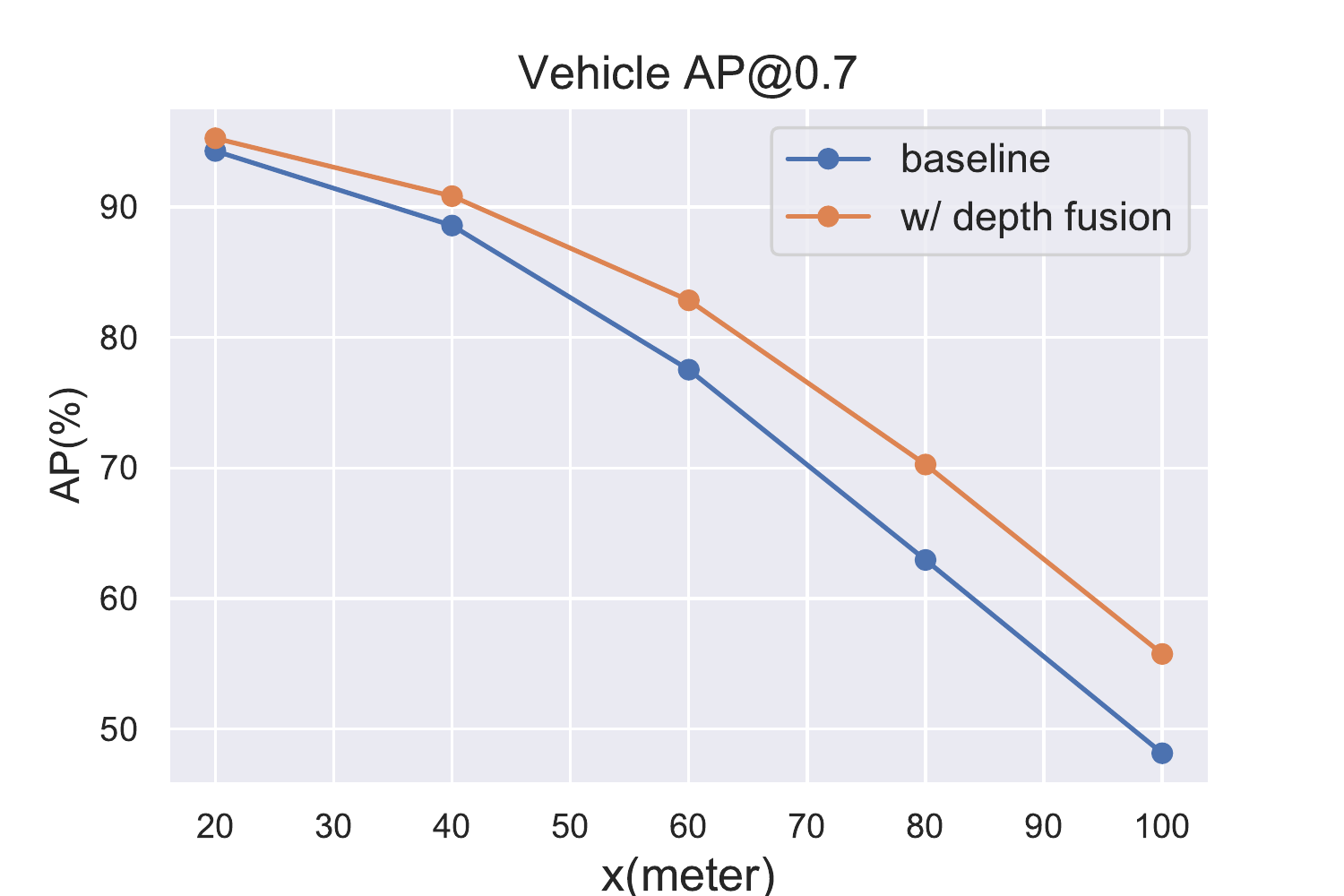} \includegraphics[width=0.3\linewidth]{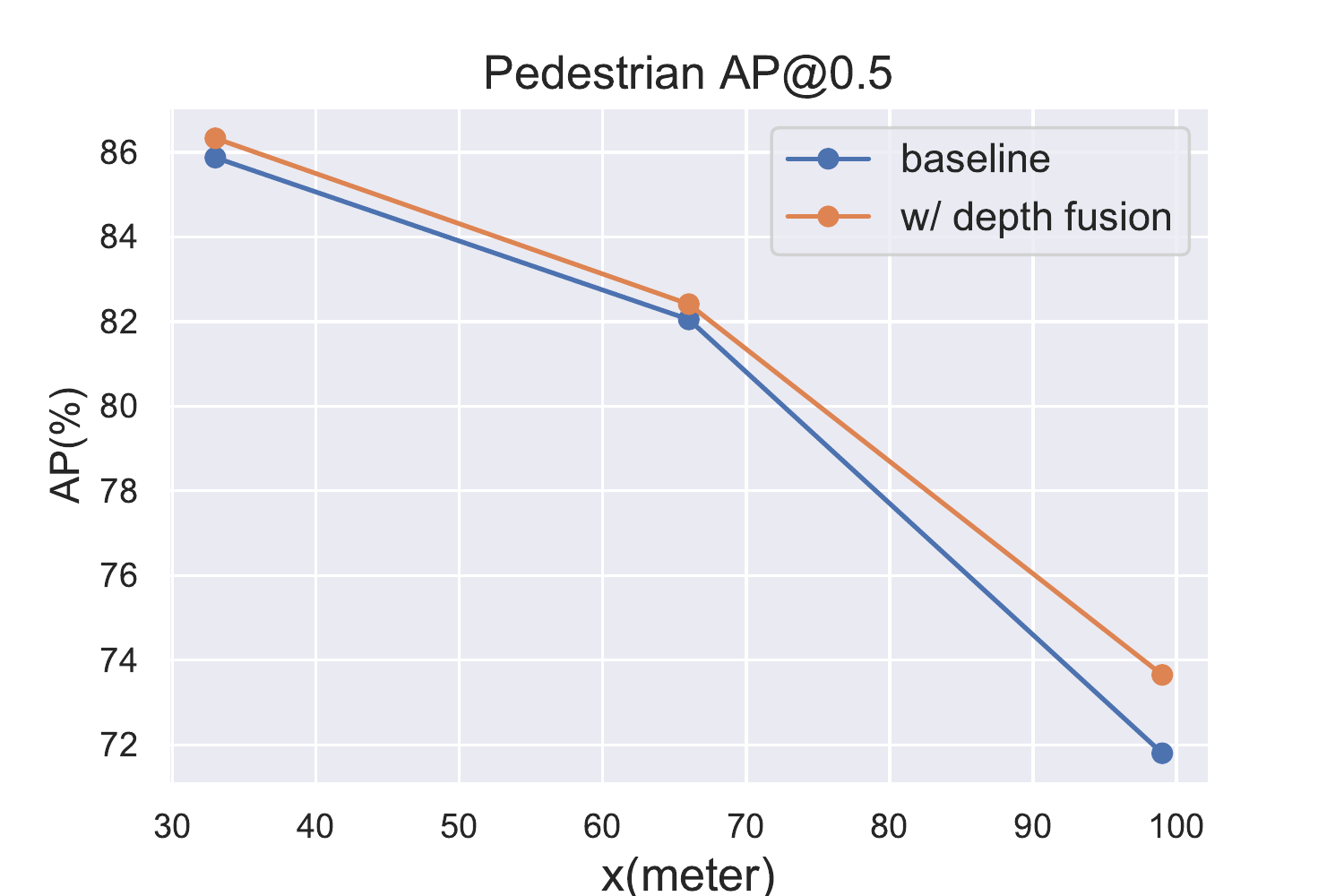} \includegraphics[width=0.3\linewidth]{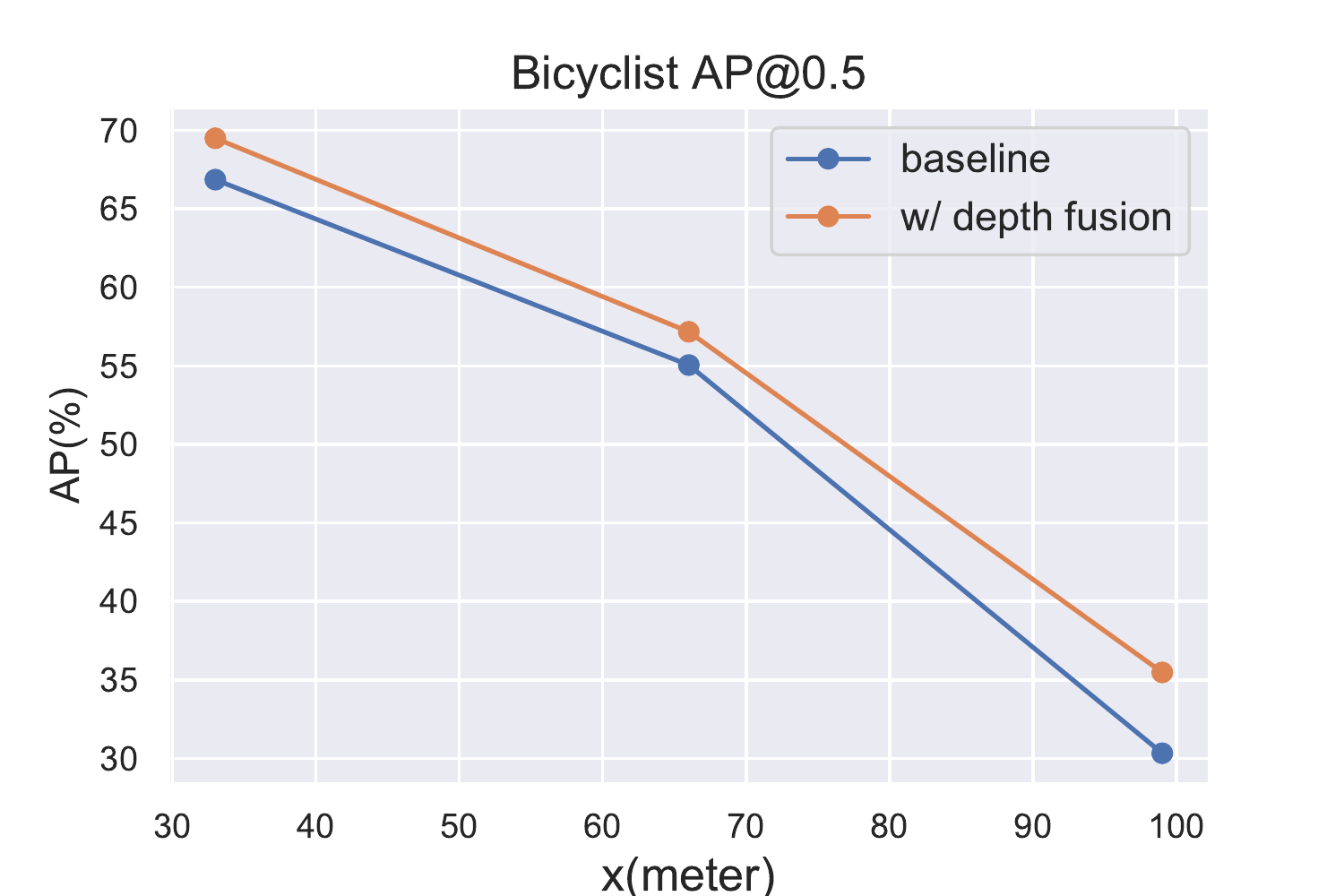}
\end{center}
   \caption{Range-wise evaluation on TOR4D BEV detection.}
\label{fig:tor4d_range}
\end{figure*}

Fig. \ref{fig:tor4d_range} shows the fine-grained evaluation results of the proposed detector on TOR4D multi-class BEV object detection at different ranges and IoU thresholds. Note that by using depth completion for dense fusion, our approach achieves larger AP gains at long range.

\begin{figure*}[t]
\begin{center}
\includegraphics[width=1.0\linewidth]{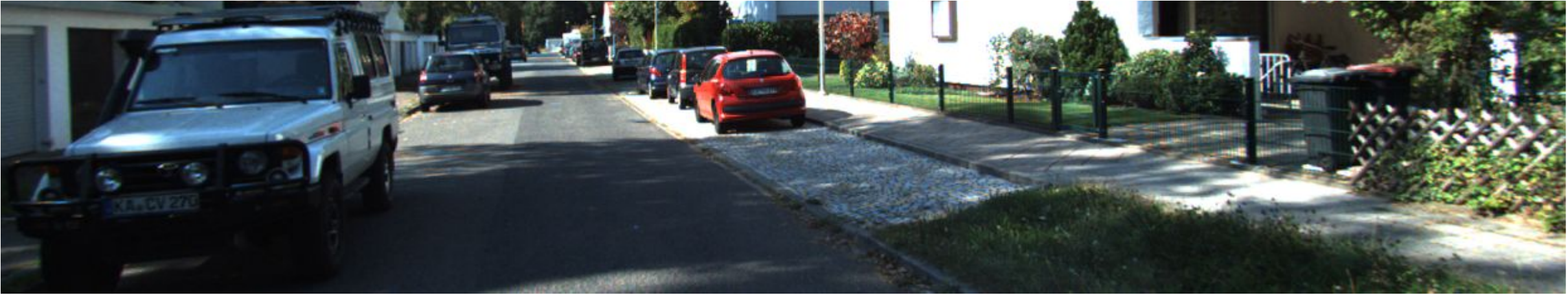}
\includegraphics[width=1.0\linewidth]{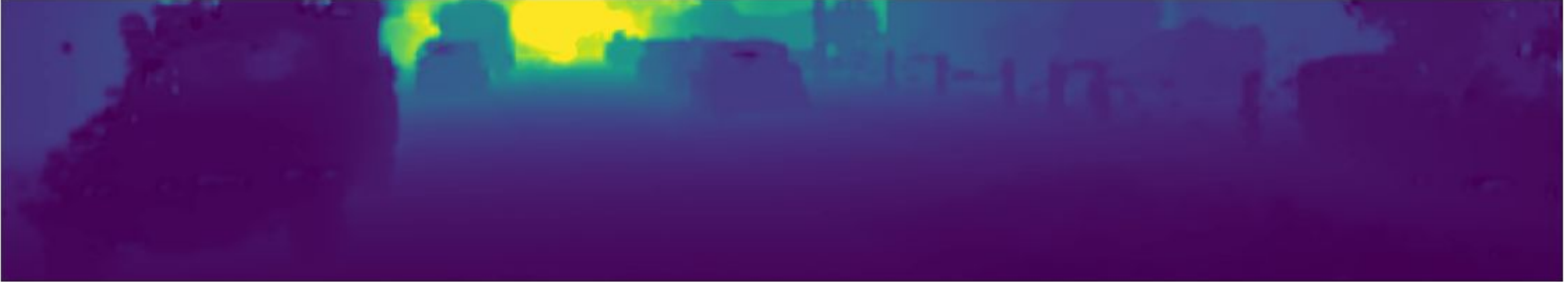}\\
\includegraphics[width=1.0\linewidth]{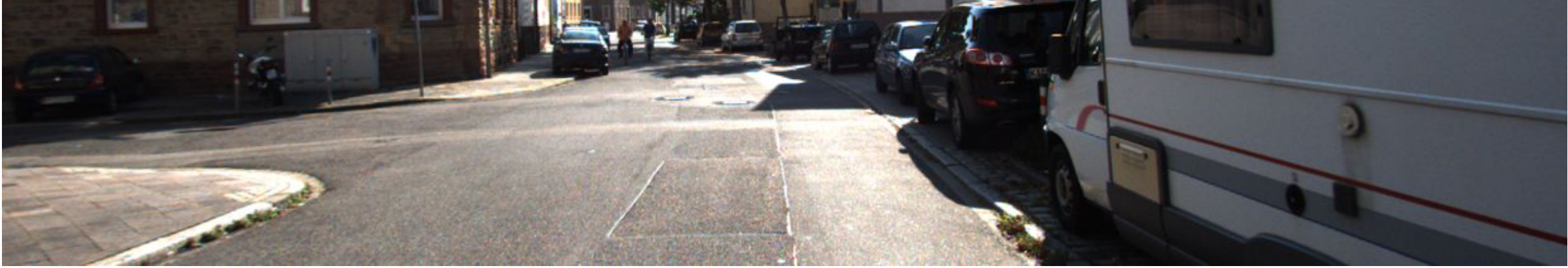}
\includegraphics[width=1.0\linewidth]{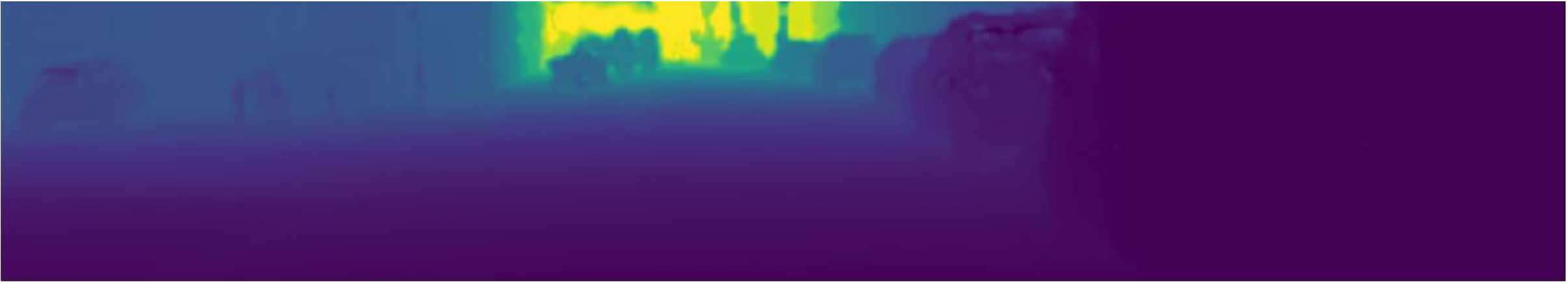}\\
\includegraphics[width=1.0\linewidth]{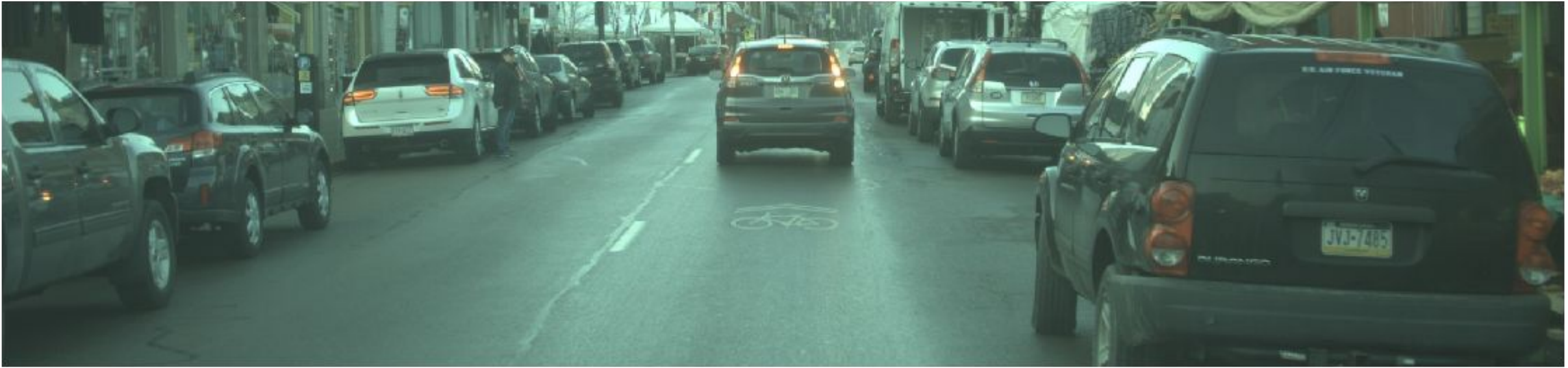}
\includegraphics[width=1.0\linewidth]{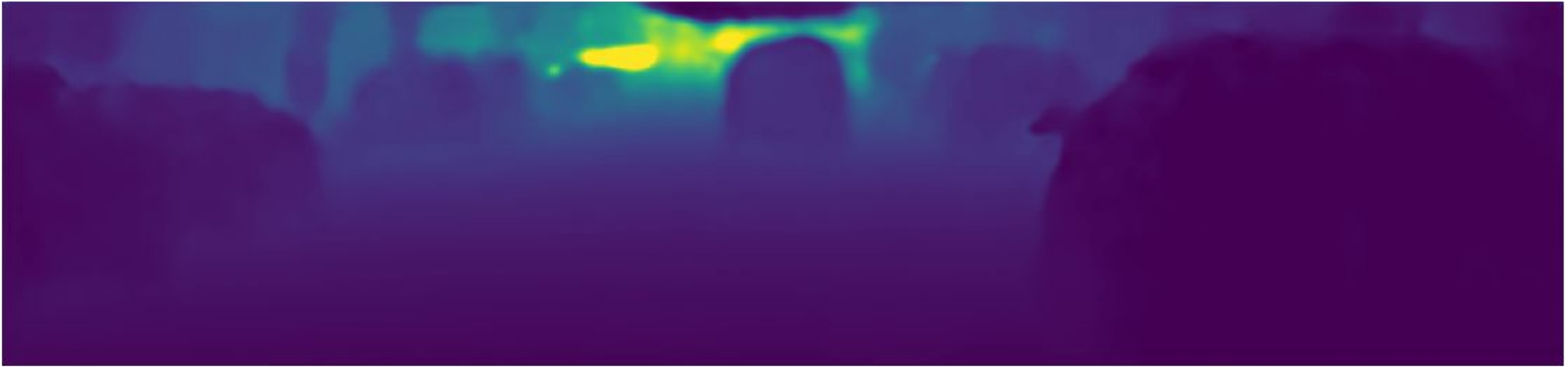}\\
\end{center}
   \caption{Qualitative results of depth completion on KITTI (first 2 examples) and TOR4D (last example).}
\label{fig:depth}
\end{figure*}

Fig. \ref{fig:depth} shows the qualitative results of depth completion on KITTI and TOR4D. Note that the camera on TOR4D has longer focal length, therefore the input depth image is more sparse. But the objects with predicted depth are also farther away, leading to more gain in long range detection.

\end{document}